\documentclass[lettersize,journal]{IEEEtran}
\usepackage{amsmath,amsfonts}
\usepackage{algorithmic}
\usepackage{algorithm}
\usepackage{array}
\usepackage[caption=false,font=normalsize,labelfont=sf,textfont=sf]{subfig}
\usepackage{textcomp}
\usepackage{stfloats}
\usepackage{url}
\usepackage{verbatim}
\usepackage{graphicx}
\usepackage{cite}
\usepackage{diagbox}
\usepackage{tabularray}
\usepackage{subcaption}
\usepackage{xcolor}
\begin{document}

\title{SeFi-CD: A Semantic First Change Detection Paradigm That Can Detect Any Change You Want}

\author{Ling Zhao, Zhenyang Huang, Dongsheng Kuang, Chengli Peng, Jun Gan, Haifeng Li,~\IEEEmembership{Member,~IEEE,}
\thanks{Ling Zhao, Zhenyang Huang, Dongsheng Kuang, Chengli Peng, and Haifeng Li are with the School of Geosciences and Info-Physics, 
Central South University, Changsha 410083, China
(e-mail: 
zhaoling@csu.edu.cn;
zhenyanghuang7@gmail.com;
215012192@csu.edu.cn;
pengcl@csu.edu.cn;
lihaifeng@csu.edu.cn).}
\thanks{Jun Gan is with China Railway Design Corporation, Tianjin, 300308, China
(e-mail:
crdcganjun@163.com).}
}

\markboth{Journal of \LaTeX\ Class Files,~Vol.~14, No.~8, August~2021}%
{Shell \MakeLowercase{\textit{et al.}}: A Sample Article Using IEEEtran.cls for IEEE Journals}


\maketitle

\begin{abstract}
The existing change detection(CD) methods can be summarized as the visual-first change detection (ViFi-CD) paradigm, which first extracts change features from visual differences and then assigns them specific semantic information. 
However, CD is essentially dependent on change regions of interest (CRoIs), meaning that the CD results are directly determined by the semantics changes of interest, making its primary image factor semantic of interest rather than visual.
The ViFi-CD paradigm can only assign specific semantics of interest to specific change features extracted from visual differences, leading to the inevitable omission of potential CRoIs and the inability to adapt to different CRoI CD tasks.
In other words, changes in other CRoIs cannot be detected by the ViFi-CD method without retraining the model or significantly modifying the method.
This paper introduces a new CD paradigm, the semantic-first CD (SeFi-CD) paradigm. 
The core idea of SeFi-CD is to first perceive the dynamic semantics of interest and then visually search for change features related to the semantics.
Based on the SeFi-CD paradigm, we designed Anything You Want Change Detection (AUWCD). Experiments on public datasets demonstrate that the AUWCD outperforms the current state-of-the-art CD methods,
achieving an average F1 score 5.01\% higher than that of these advanced supervised baselines on the SECOND dataset, with a maximum increase of 13.17\%.
The proposed SeFi-CD offers a novel CD perspective and approach.
\end{abstract}


\section{Introduction}

\begin{figure}[htbp]
  \centering
  \includegraphics[width=3.7in]{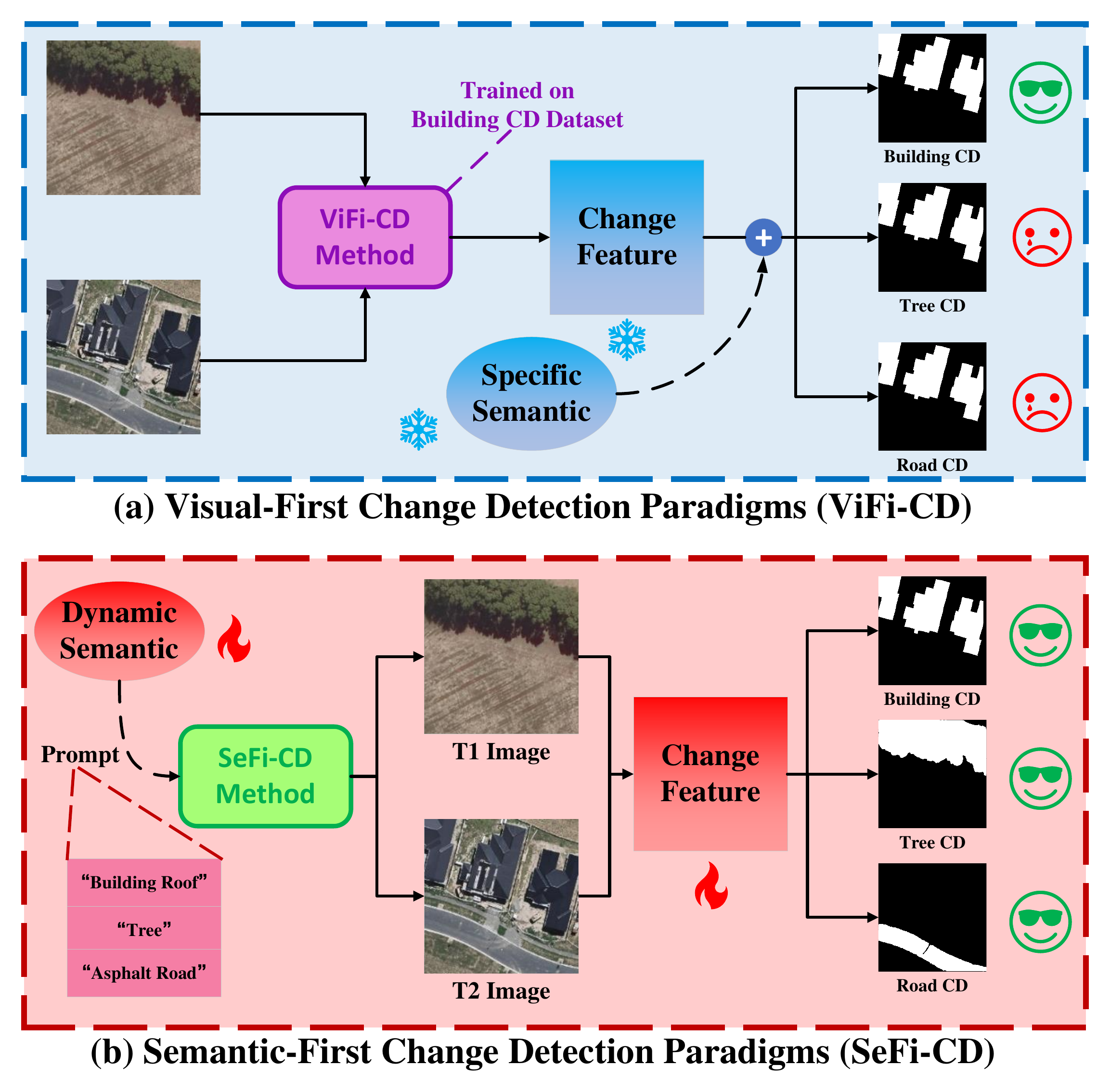}
  \caption{Comparison diagram of the ViFi-CD and SeFi-CD paradigms.
  The ViFi-CD paradigm first extracts specific change features visually and then assigns them specific semantic information,
  leading to methods that can only detect specific types of CRoIs. 
  In contrast, the SeFi-CD paradigm first determines the semantic information of changes based on user prompts and then searches for corresponding and dynamic change features in the image, 
  allowing it to detect CRoIs of any semantic type.}
  \label{fig2} 
  \end{figure}

Change detection (CD) is an important remote sensing task and can be used in tasks such as disaster assessment\cite{a_split-based_approach_to_unsupervised_change_detection_in_large-size_multitemporal_images:_application_to_tsunami-damage_assessment},
environmental monitoring\cite{digital_change_detection_methods_in_natural_ecosystem_monitoring:_a_review},
land management\cite{corine_land_cover_change_detection_in_europe_(case_studies_of_the_netherlands_and_slovakia)} and
urban change analysis\cite{land_use/land_cover_change_detection_and_urban_sprawl_analysis}.
The main purpose of the CD task is to detect changes in a specific target; however, not all changes in images of different phases since any location can change over time.
The target we focus on can change according to the needs of subsequent tasks. For example, in practical applications, when we wish to investigate changes in living space, we focus on changes in buildings; when we want to monitor changes in traffic,
we focus on changes in vehicles. 
The regions we focus on can be called \textbf{change regions of interest (CRoIs)}, while other targets can be called \textbf{change regions of uninterest (CRoUI)}.
Since the change maps of different CRoIs differ, the result of CD is determined by the semantics of interest, and the nature of the interest is considered semantic information.
Therefore, in this paper, we propose that \textbf{the semantic of interest is the first image factor in CD tasks.}.\par

In addition to semantics, visualizations are also an important image factor in CD tasks. 
The current CD paradigm can be summarized as the visual first CD (ViFi-CD) paradigm, as shown in Fig.\ref{fig2}(a),
which first extracts specific change features from visual changes and subsequently gives the change features specific semantics in terms of fixation interests. 
There are two main problems with this paradigm, which we analyze in detail in Section 2.
These two problems can be roughly described as follows:
\begin{enumerate}
  \item{
  Unsupervised ViFi-CD methods tend to easily interpret visual differences as changes, but visual differences sometimes do not imply changes in the semantic content of interest.
  }
  \item{
  Deep learning-led supervised Learning ViFi-CD methods tend to result in models that can only extract specific visual features from specific CRoI and detect those specific CRoIs after training on them while ignoring other potential CRoIs.
  Without additional training, the model can not detect different CRoIs. However, conducting additional training on the model requires reannotating the training data, which is a very costly process.
  }
\end{enumerate}
\par
To solve the above two problems of the ViFi-CD paradigm, the SeFi-CD paradigm is proposed. As Fig.\ref{fig2}(b) shows, we can use any dynamic CRoI semantics to
prompt the SeFi-CD model, informing the model about the changes we are interested in, allowing the model to understand our needs, and then actively searching for the corresponding change
features. The paradigm can search for dynamic change features based on dynamic semantics and can be adapted to different CRoI CD tasks.\par

To our knowledge, no CD method exists for the SeFi-CD paradigm.
Therefore, we propose the first CD framework \textbf{AUWCD} for the SeFi-CD paradigm in this paper.
Several change scenarios are detected in the CD task due to the diversity in task requirements.
This framework can detect corresponding changes according to different CRoI semantic prompts.
The proposal of the AUWCD is mainly motivated by the following three factors:
\begin{enumerate}
  \item{\textbf{Human describability of CRoI semantics}.
  Different tasks require different CRoI semantics. First, we must be able to obtain an accurate representation of different CRoI semantics. 
  Language, as a commonly used means of communication for human beings, covers the linguistic expression of most features; therefore, language can be used as a carrier of different CRoI semantics.}
  \item{\textbf{The semantic-first principle}.
  In recent years, "prompt engineering" has gradually replaced the previous paradigm of "pretraining fine-tuning" for downstream tasks\cite{pre-train__prompt__and_predict:_a_systematic_survey_of_prompting_methods_in_natural_language_processing}, which allows the model to perform on large quantities of raw data.
  Pretraining is performed on the model, and by defining a new prompt function, the model can perform few-shot or even zero-shot learning and adapt to new scenarios with little or no labeled data.
  Therefore, we can use the CRoI concept described in language as a semantic prompt and prioritize prompts to guide the model to extract corresponding change features;}
  \item{\textbf{The ability of the model to understand language}.
  Since we choose human language to express CRoI semantics, we also need a model that can understand the language-expressed CRoI semantics.
  Here, we introduce a large multimodal visual language model to solve this problem.}
\end{enumerate}
\par In summary, our contributions are as follows:
\begin{itemize}
  \item{We summarize the current CD methods as a visual-first CD paradigm and provide a detailed analysis of the issues associated with this paradigm.}
  \item{In response to the issues with ViFi-CD, we propose a new CD paradigm called the SeFi-CD paradigm. 
        This paradigm dynamically locates regions of interest based on user-provided CRoI text prompts. 
        Additionally, we design AUWCD, the first CD framework under the SeFi-CD paradigm.}
  \item{We compare the AUWCD with current state-of-the-art CD methods on two public CD datasets.
  The experimental results show that when the CRoI is modified according to task requirements without additional training,
  the AUWCD has strong flexibility and better detection accuracy.
  On the second dataset, the F1 score exceeds that of these advanced supervised baselines by an average of 5.01\%, with the highest exceeding 13.17\%.}
  \end{itemize}
\par The remainder of this article is organized as follows. 
Section 2 provides a detailed analysis of the issues present in the ViFi-CD paradigm.
Section 3 reviews the related works. 
Section 4 describes the proposed method in detail. 
To evaluate our method, experiments are designed in Section 5. 
Finally, Section 6 summarizes this article and then discusses future research directions based on the experimental results.\par

\section{Comparative Analysis of ViFi-CD and SeFi-CD}
In this section, we conduct a detailed analysis of the issues associated with the current ViFi-CD methods.
We also compare ViFi-CD methods with SeFi-CD methods to provide a deeper understanding of SeFi-CD.\par

In the Introduction section, we discussed two crucial image factors in CD: \textbf{semantic} and \textbf{visual}.
To comprehensively analyze the issues with ViFi-CD, 
we categorized the changes into four scenarios based on both the semantic and visual perspectives, as shown in Table \ref{tab:table1}:
(a) \textbf{both semantic and visual change}, (b) \textbf{semantic change but no visual change}, (c) \textbf{no change in semantic but visual change}, and (d) \textbf{no semantic or visual changes}.
It should be noted that "semantic" here is the change in an interesting semantic because the change process usually involves a change in some, not all semantics.
Moreover, in reality, there is no absolute "unchange" visually due to lighting, imaging, etc. 
Therefore, visual change and unchange can be understood as visually "obvious differences" and "subtle differences", respectively.
Cases (a), (b) and (c) are usually CRoIs in various CD tasks, and case (d) is a CRoUI in most cases.
\begin{table*}[htbp]
  \caption{Four situations in which changes are divided between "semantic" and "visual" perspectives.\label{tab:table1}}
  \centering
  \begin{tabular}{c|c|c} 
  \hline
  & \textbf{Visual Change} & \textbf{Visual Unchange}  \\ 
  \hline
  \begin{tabular}[c]{@{}c@{}}\textbf{Semantic}\\\textbf{Change}\end{tabular}  & \begin{minipage}{.4\textwidth} \includegraphics[width=2.8in]{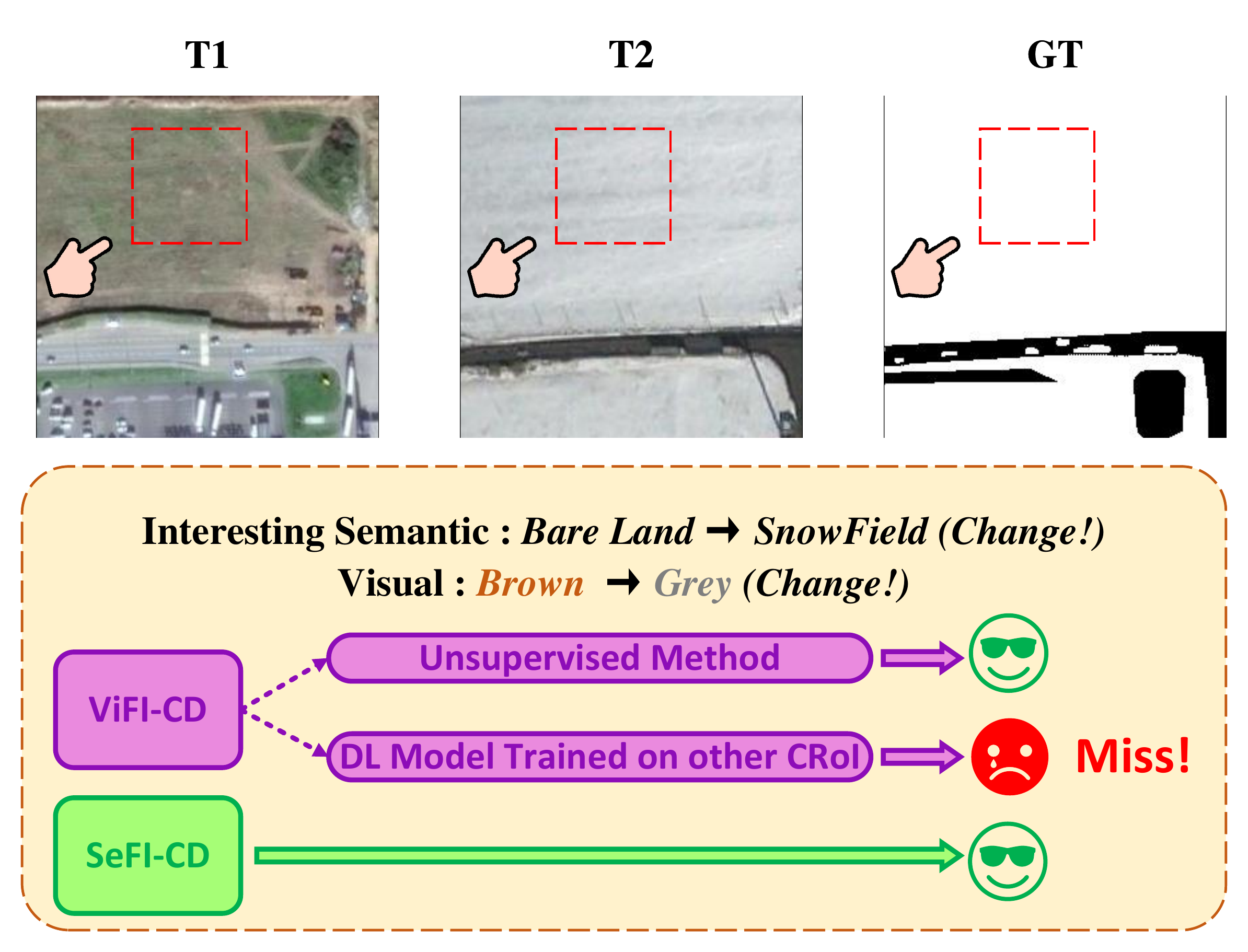} \begin{center} (a) \end{center} \end{minipage}  & \begin{minipage}{.4\textwidth} \includegraphics[width=2.8in]{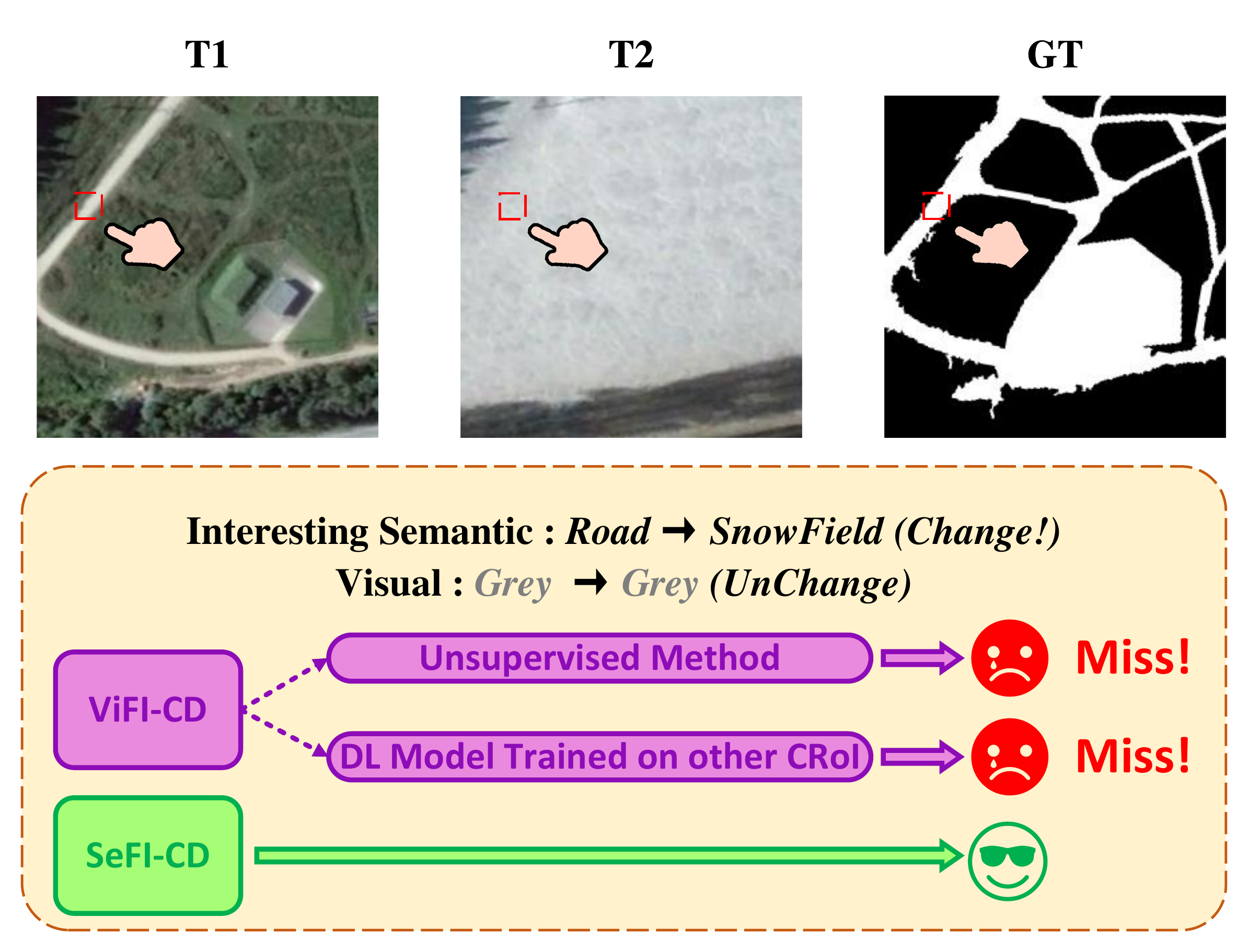}  \begin{center} (b) \end{center} \end{minipage}     \\ 

  \hline
  \begin{tabular}[c]{@{}c@{}}\textbf{Semantic}\\\textbf{Unchange}\end{tabular}  & \begin{minipage}{.4\textwidth} \includegraphics[width=2.8in]{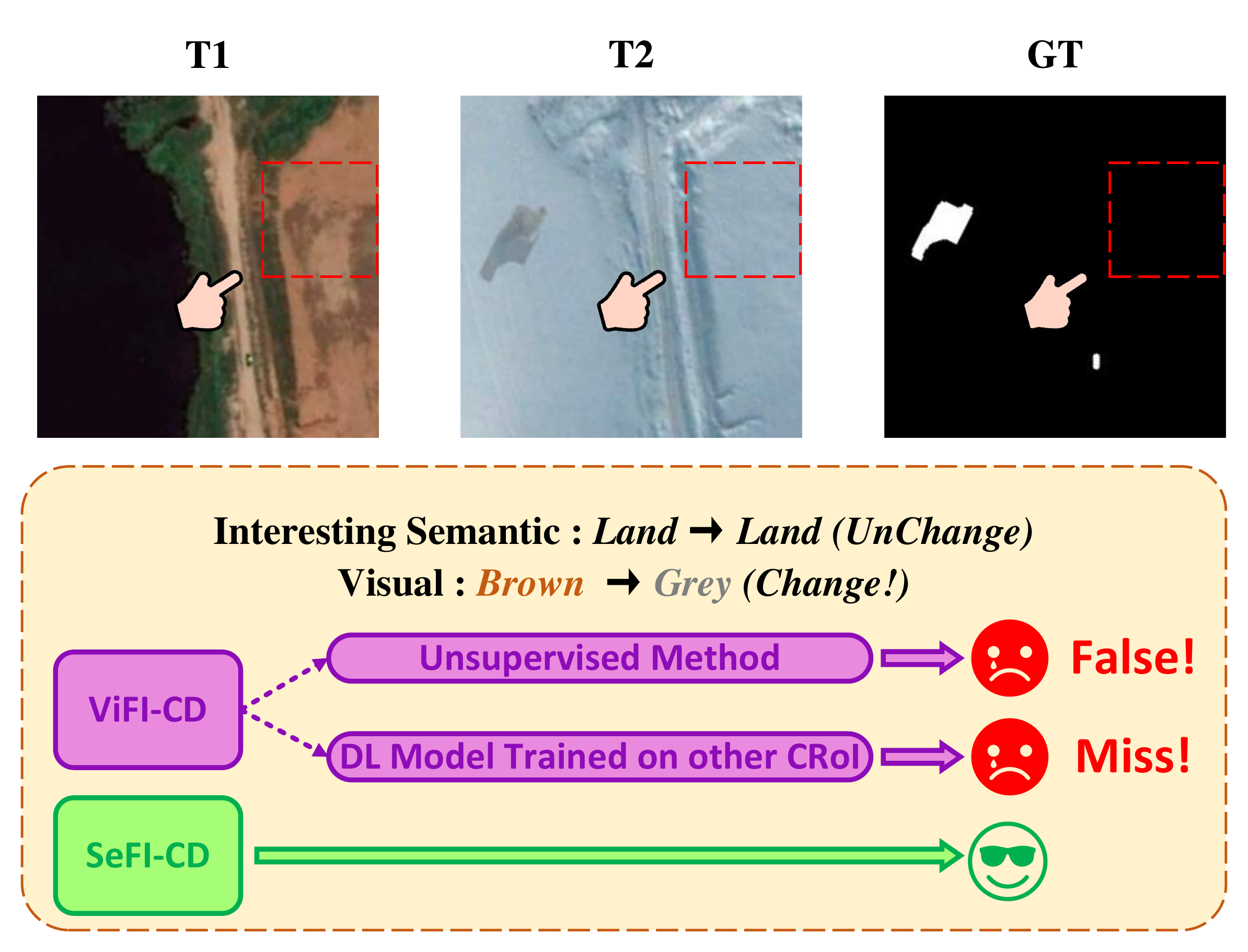} \begin{center} (c) \end{center} \end{minipage}   & \begin{minipage}{.4\textwidth} \includegraphics[width=2.8in]{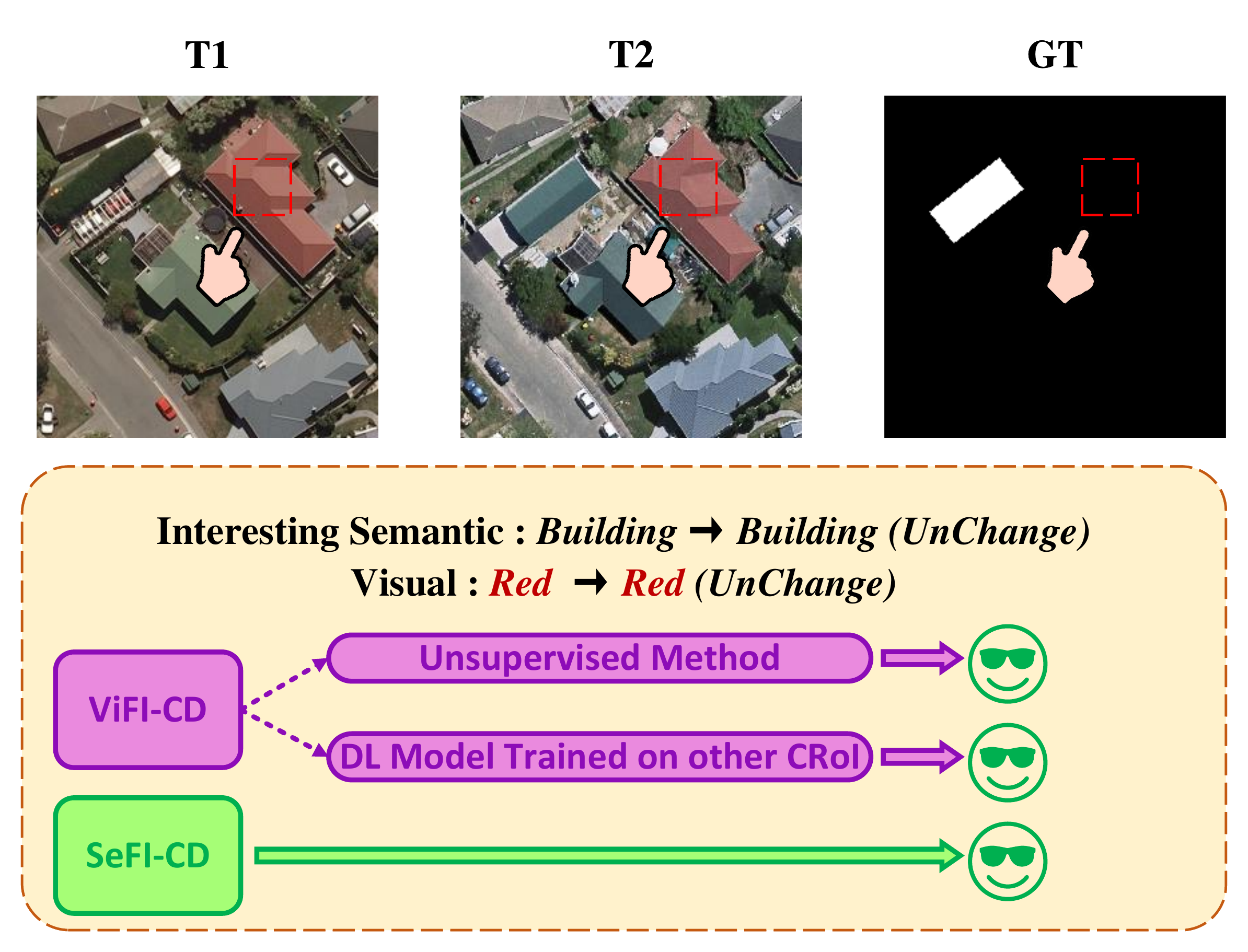} \begin{center} (d) \end{center} \end{minipage}     \\
  \hline
  \end{tabular}
  \end{table*}
\par
The methods within the ViFi-CD paradigm can be categorized into \textbf{unsupervised methods} and \textbf{deep learning-led Supervised Learning Methods}.
In the following sections, we conduct an issue analysis on these two types of methods separately and compare them with SeFi-CD.
\subsection{Unsupervised CD Methods}
Before the popularization of deep learning technology, unsupervised methods were the mainstream methods for CD tasks.
To date, researchers have proposed various unsupervised CD methods,
which can be roughly divided into \textbf{algebra-based} methods and \textbf{transform-based} methods\cite{change_detection_techniques_for_remote_sensing_applications:_a_survey}.
The \textbf{algebra-based} methods obtain the difference map by applying mathematical operations to each image pixel and then obtain the final change map through operations
\cite{
change_detection_techniques_for_remote_sensing_applications:_a_survey,
procedures_for_change_detection_using_landsat_digital_data,
adaptive_change_detection_with_significance_test,
sequential_spectral_change_vector_analysis_for_iteratively_discovering_and_detecting_multiple_changes_in_hyperspectral_images,
detecting_changes_between_optical_images_of_different_spatial_and_spectral_resolutions:_a_fusion-based_approach,
a_change_detection_measure_based_on_a_likelihood_ratio_and_statistical_properties_of_sar_intensity_images,
a_generalized_likelihood_ratio_test_for_coherent_change_detection_in_polarimetric_sar}.
The \textbf{transform-based} CD method first transforms the image pixels and then processes the transformed feature map to obtain the change map
  \cite{
design_and_implementation_of_an_expert_system_for_updating_thematic_maps_using_satellite_imagery_(case_study:_changes_of_lake_urmia),
fast_detection_of_significantly_transformed_areas_due_to_illegal_waste_burial_with_a_procedure_applicable_to_landsat_images,
an_effective_hybrid_classification_approach_using_tasseled_cap_transformation_(tct)_for_improving_classification_of_land_use/land_cover_(lu/lc)_in_semi-arid_region:_a_case_study_of_morva-hadaf_watershed__gujarat,
an_approach_for_unsupervised_change_detection_in_multitemporal_vhr_images_acquired_by_different_multispectral_sensors,
applying_the_chi-square_transformation_and_automatic_secant_thresholding_to_landsat_imagery_as_unsupervised_change_detection_methods,
an_approach_based_on_discrete_wavelet_transform_to_unsupervised_change_detection_in_multispectral_images,
a_deep_convolutional_coupling_network_for_change_detection_based_on_heterogeneous_optical_and_radar_images,
learning_multiscale_deep_features_for_high-resolution_satellite_image_scene_classification,
a_fully_learnable_context-driven_object-based_model_for_mapping_land_cover_using_multi-view_data_from_unmanned_aircraft_systems,
efficient_patch-wise_semantic_segmentation_for_large-scale_remote_sensing_images,
change_detection_in_heterogenous_remote_sensing_images_via_homogeneous_pixel_transformation
}.\par
These methods focus on automatically recognizing visual differences between images through algorithms to discover change regions and then identifying regions that change the human judgment of semantic information\cite{change_detection_techniques_for_remote_sensing_applications:_a_survey}.
This process usually involves comparing two temporal images and finding visual change features (e.g., brightness, color, and texture)\cite{sequential_spectral_change_vector_analysis_for_iteratively_discovering_and_detecting_multiple_changes_in_hyperspectral_images}.
This method is effective when it involves obvious visual changes, such as the addition or destruction of buildings.
However, these methods suffer from two critical issues:
\begin{itemize}
  \item{\textbf{Missing regions with minor visual differences but significant semantic disparities}.
        As Table \ref{tab:table1}(b) shows, this region exhibits minor visual differences, appearing as gray-white areas, but undergoes a semantic change from "Road" to "Snow Field".
        Unsupervised methods typically rely only on visual differences for determination, often classifying such regions as unchanged, which can lead to missed changes.}
  \item{\textbf{False positives for CRoUI}. As Table \ref{tab:table1}(c) shows, this region exhibits significant visual differences; however, when the semantic interest is "Land", no change exists in this area.
        However, unsupervised methods may incorrectly identify these obvious visual differences in the region as changes, leading to false positives in CD.}
  \end{itemize}
\par
For the first issue, SeFi-CD prioritizes semantic changes based on the provided semantic prompt, enabling the identification of change features in the image that are relevant to the specified semantic interest.
For the second issue, SeFi-CD focuses only on change features relevant to the specified semantic interest, disregarding changes that are irrelevant to the specified semantics.

\subsection{Deep Learning-Led Supervised Learning Methods}
With its powerful representation capabilities, deep learning technology has become the mainstream method for current CD tasks and has achieved surprising results.
Deep learning algorithms, which aim to hierarchically learn representative and discriminative features from datasets, have received widespread attention from the global earth science and remote sensing communities.
Researchers have recently proposed various deep network models for CD tasks.
These deep networks include convolutional neural networks (CNNs), transformers, generative adversarial networks (GANs), and recurrent neural networks (RNNs)
\cite{comprehensive_review_and_meta_analysis}. 
The CNN and transformer structures are the main model structures in this type of CD method.
A CNN is a special form of neural network used to process data with a known grid-like representation, such as image data, which can be thought of as a two-dimensional grid of pixels
\cite{
a_survey_on_deep_learning-based_change_detection_from_high-resolution_remote_sensing_images,
fully_convolutional_networks_for_semantic_segmentation,
a_survey_on_deep_learning-based_change_detection_from_high-resolution_remote_sensing_images,
fully_convolutional_siamese_networks_for_change_detection,
learning_to_measure_change:_fully_convolutional_siamese_metric_networks_for_scene_change_detection,
fully_convolutional_siamese_networks_based_change_detection_for_optical_aerial_images_with_focal_contrastive_loss,
dasnet,
a_deeply_supervised_attention_metric-based_network_and_an_open_aerial_image_dataset_for_remote_sensing_change_detection,
ads-net:an_attention-based_deeply_supervised_network_for_remote_sensing_image_change_detection,
a_spatial-temporal_attention-based_method_and_a_new_dataset_for_remote_sensing_image_change_detection,
snunet-cd}.
The 
transformer\cite{transformer}
was first introduced in 2017 and was originally used for sequence-to-sequence learning.
Transformers can easily provide long-range dependency modeling; therefore, they have been widely used in natural language processing (NLP) and are starting to show promise in computer vision and CD.
\cite{
a_survey_on_deep_learning-based_change_detection_from_high-resolution_remote_sensing_images,
a_transformer-based_network_for_change_detection_in_remote_sensing_using_multiscale_difference-enhancement,
bit,
transunetcd,
transcd,
swinsunet}.
\par

This type of method detects change regions by training CD models using labeled datasets\cite{fully_convolutional_change_detection_framework_with_generative_adversarial_network_for_unsupervised__weakly_supervised_and_regional_supervised_change_detection},
which can somewhat overcome the shortcomings of unsupervised methods, as they can learn complex patterns of specific change types\cite{deep_learning}.
However, the effectiveness of this method is limited by the type and number of labels in the dataset. 

Moreover, As Table \ref{tab:table1} shows, a more severe issue with these methods is that once trained on specific CRoIs,
the model can only extract visual features specific to those specific CRoIs and can not detect other CRoIs, leading to missed detections of other CRoIs.
For example, if we train a deep learning model for road CD using a road CD dataset,
the model becomes ineffective when tasked with building CD, resulting in significant missed detections of building changes.

Although this problem can be solved by retraining the model by relabelling the data, labeling the training samples is quite time-consuming and laborious\cite{fully_convolutional_change_detection_framework_with_generative_adversarial_network_for_unsupervised__weakly_supervised_and_regional_supervised_change_detection}.
Later, researchers gradually recognized the importance of semantics in CD tasks.
For example, Dong et al. proposed ChangeCLIP\cite{changeclip}, the first work to apply a multimodal vision-language approach to CD tasks. 
ChangeCLIP introduces rich semantic information using pre-trained CLIP\cite{clip}, providing semantic guidance for model training and enhancing the performance compared to using a single visual modality alone.
However, this method still starts with visual features and the model cannot dynamically search for change features related to the user's specified semantic interests.
Therefore, this method still faces the aforementioned issues.
\par
As illustrated in Figure 1, SeFi-CD can dynamically accept user-provided CRoI prompts,
enabling it to comprehend diverse task requirements and actively search for corresponding change features.
This approach resolves the issues associated with the deep learning-based supervised methods prevalent in ViFi-CD.

\section{Related Works}
In this section, to better explain the contents of VLMs and FSMs in the AUWCD, we summarize the development of vison-language pretrained models and foundation segment models.
The details are as follows: \par

\subsection{Vision-Language Pretrained Models}
In recent years, with the continued interest in the multimodal field and the rapid development of self-supervised learning, a series of pretrained visual language models represented by
CLIP\cite{clip} and ALIGN\cite{align} have been created.
These models usually consist of modality-specific (image and text) encoders capable of generating embeddings for various modalities.
These models are usually pretrained on billion-scale image and text datasets, and contrastive learning methods are used to maximize the alignment of image-text positive sample pairs by randomly sampling image-text pairs.
A direct application of this type of model is zero-shot image-text retrieval or zero-shot classification through text
prompts\cite{clip},
which explore shared or mixed architectures between image and text modalities and implement additional features such as zero-shot visual question answering (VQA) and subtitles.
There are also several methods,
such as locked image tuning(LiT) (Zhai et al., 2022)\cite{lit},
aligning pretrained encoders(APE) (Rosenfeld et al., 2022)\cite{ape},
and  bootstrapping language-image pretraining(BLIP-2) (Li et al., 2023a)\cite{blip-2}, 
that uses a pretrained single-modal model to reduce the training cost of 
CLIP-like models\cite{sam-clip}.\par

While large pretrained visual language models have shown amazing capabilities, researchers have also conducted in-depth discussions on the interpretability of such models.
CLIP surgery\cite{clip_surgery} refers to the concept of MaskCLIP\cite{extract_free_dense_labels_from_clip}
in open vocabulary semantic segmentation.
This task requires neither extra supervision nor additional training. In contrast to MaskCLIP,
CLIP surgery aims to improve the interpretability of CLIP\cite{clip}, while the goal of MaskCLIP is to extract dense predictions for segmentation; CLIP surgery does not delete self-attention in MaskCLIP but rather proposes v-v self-attention.
The original attention module is transformed; in addition, CLIP surgery introduces dual paths to merge multiple v-v self-attention mechanisms instead of only modifying the last layer, as in MaskCLIP. CLIP surgery is an interpretable method designed for CLIP models.
It uses text prompts to generate accurate explanation maps, which can highlight areas that are similar to text concepts.\par

\subsection{Foundation Segment Models}
The segment anything model(SAM)\cite{sam}, proposed by Meta in 2023, has made significant progress in breaking segmentation boundaries and greatly promoted the development of basic computer vision models.
SAM also refers to using large model prompts for powerful zero-shot learning and can generate high-quality segmentation masks based on various visual segmentation prompts.\par

Since then, many scholars have conducted research to explore the capabilities of SAM and apply it to various tasks
\cite{a_comprehensive_survey_on_segment_anything_model_for_vision_and_beyond,
can_sam_segment_anything?_when_sam_meets_camouflaged_object_detection,
generalist_vision_foundation_models_for_medical_imaging:_a_case_study_of_segment_anything_model_on_zero-shot_medical_segmentation
}.
In addition, some scholars are working on various visual prompts to improve the versatility of SAM. 
\cite{a_comprehensive_survey_on_segment_anything_model_for_vision_and_beyond,segment_everything_everywhere_all_at_once}
introduced more diverse prompts than SAM does and proposed a more general segmentation system, SEEM, which includes visual prompts (points, boxes, graffiti, masks) and text and quote hints (reference areas of another image).
As the authors claimed, the unified prompting scheme introduced in the 
SEEM\cite{segment_everything_everywhere_all_at_once} can encode different prompts into a joint visual-semantic space to produce strong zero-shot generalization capabilities.
These basic visual models can generate high-quality segmentation masks. 
Here, we regard these foundation visual models, which are designed for semantic segmentation tasks, as the foundation segment models (FSMs).
\par

\section{Method}
In this section, to better understand the working details of the AUWCD, we provide a detailed introduction to the Semantic Align Module, the CRoI Segment Module, and the CD Module, along with the correlations between these three modules.\par
\begin{figure*}[htbp]
  \centering
  \includegraphics[width=7in]{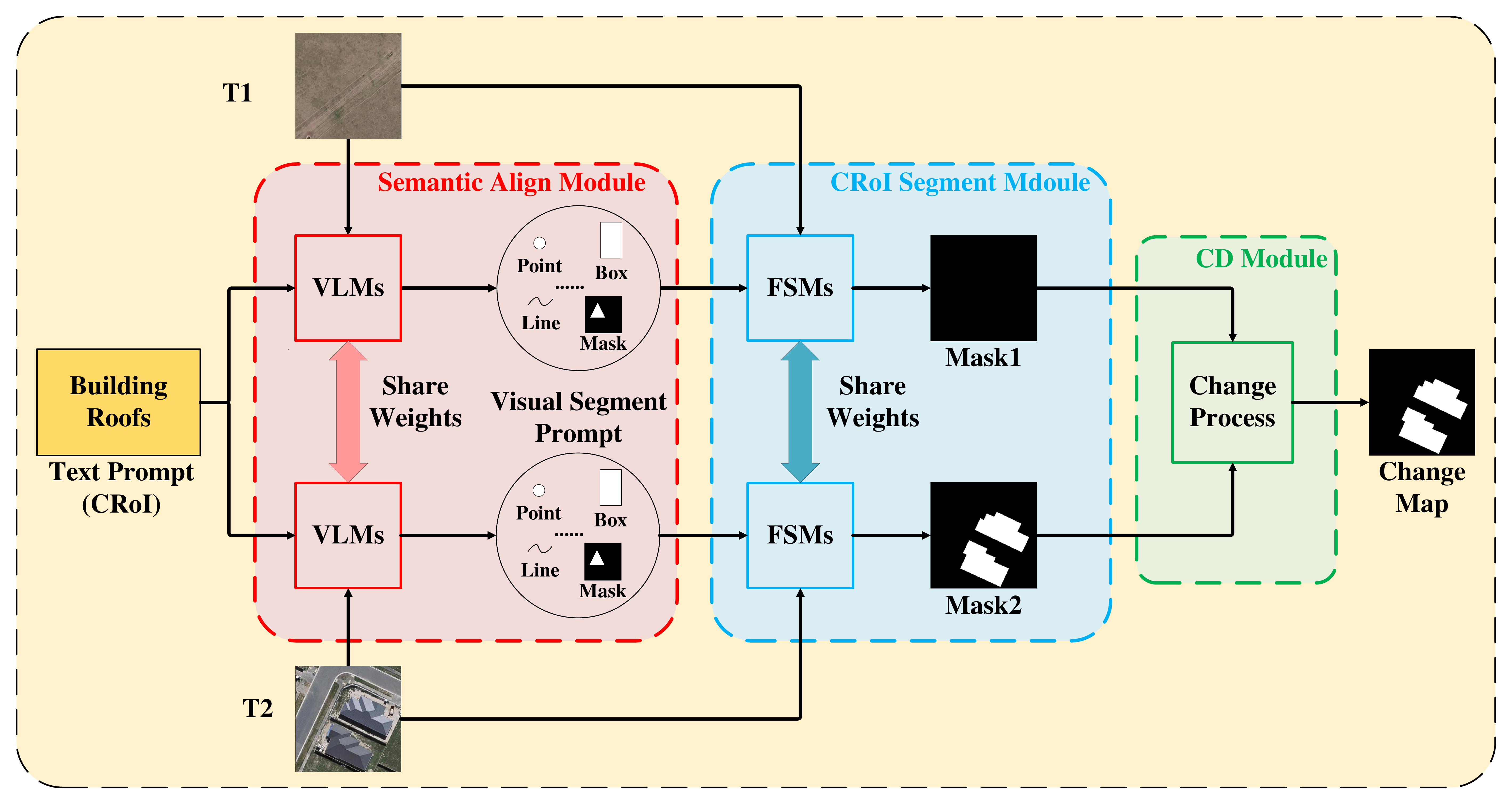}
  \caption{Overall AUWCD framework. 
  First, the Semantic Align Module models the task of interest as text prompts, obtaining visual prompts from visual language models (VLMs) based on these text prompts to dynamically perceive the semantic information of changes of interest. 
  Second, the CRoI Segment Module, based on the generated visual prompts, extracts the fine masks of CRoIs in images of different temporal phases through FSMs.
  Finally, the CD Module compares the segmentation results of images from different temporal phases to ultimately derive the CD results.}
  \label{fig3}
  \end{figure*}

Fig.\ref{fig3} shows the overall framework structure of the AUWCD. The details are as follows:\par
\subsection{Semantic Align Module}
The main function of this module is to align the feature space of the multitemporal image with the semantic space of the CRoI text prompt.
Visual prompts are generated for areas with a higher degree of alignment based on the alignment degree between the image feature space and the CRoI semantic space.
The CRoI segment module can be utilized to segment the CRoIs in images. 
VLMs are the core component of this module and usually consist of an image encoder for storing rich knowledge and a text encoder for selecting the CRoI we intend to use.
Specifically, the image encoder stores rich knowledge about more possible CRoIs and has a strong ability to distinguish their features clearly.
In other words, it can generate similar representations for CRoIs of the same type and different representations for CRoIs of different types.
We used 
CLIP Surgery\cite{clip_surgery}
to generate point prompts as an example to explain this module in detail.\par
\begin{figure}[htbp]
  \centering
  \includegraphics[width=3.4in]{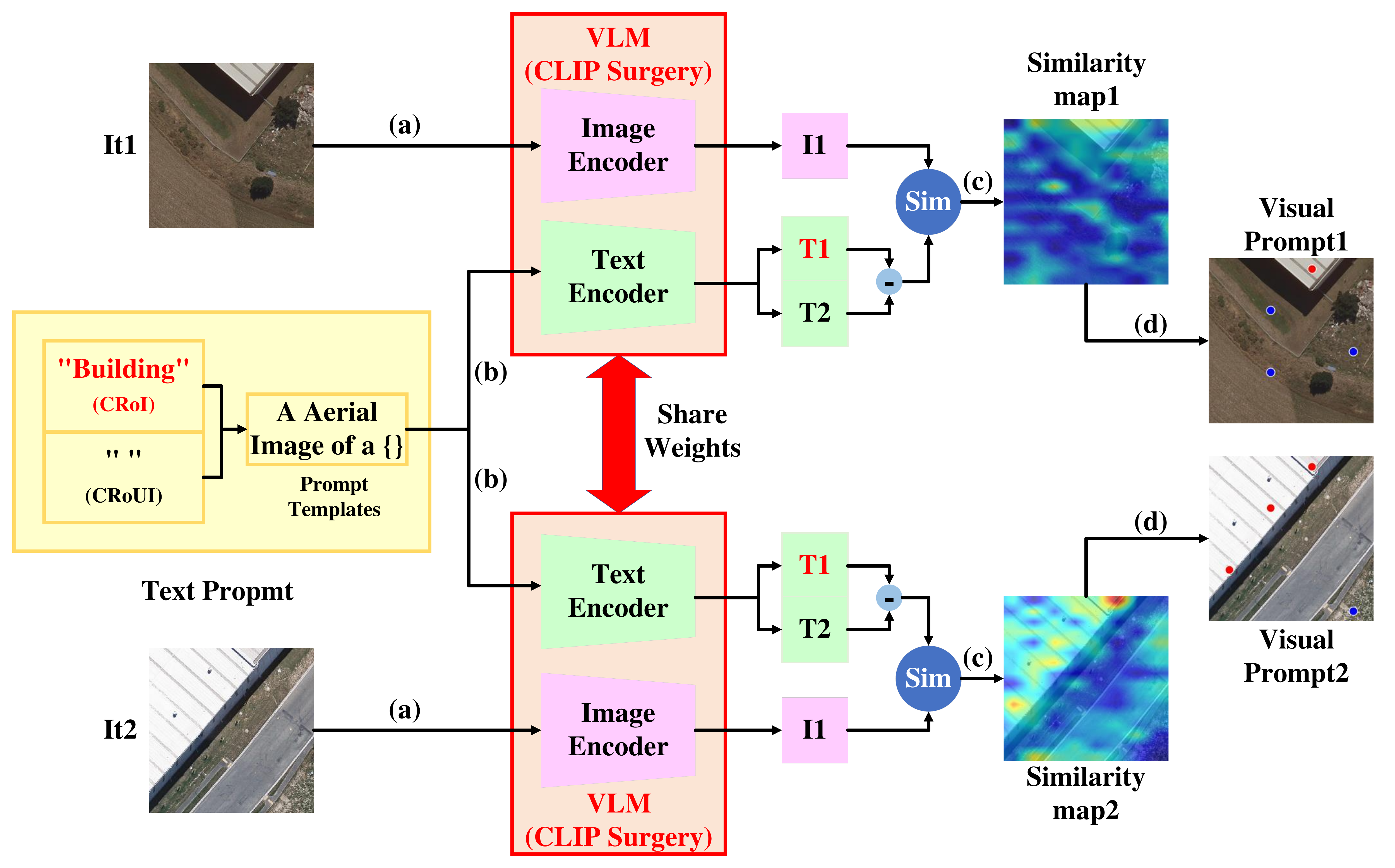}
  \caption{
    Illustration of the Semantic Align Module (CLIP surgery as an example).
    (a) shows the image encoding process.
    (b) shows the text encoding process.
    (c) shows similarity map generation.
    (d) shows point prompt generation based on similarity maps.}
  \label{fig4}
  \end{figure}

The overall process of this module is shown in Fig.\ref{fig4}. 
We describe the four processes of \textbf{image encoding}, \textbf{text encoding}, \textbf{similarity map generation}, and \textbf{point prompt generation based on similarity maps}:\par

\paragraph{Image Encoding}
The dual-temporal images $\boldsymbol{It1}$ and $\boldsymbol{It2}$ obtain the text encodings $\boldsymbol{I_1}$ and $\boldsymbol{I_2}$ of the two images through the image encoder.
The two image encoders used here are from CLIP Surgery, and their weights are shared.\par

\paragraph{Text Encoding}
Both CRoI and CRoUI are represented by text.
We can specify CRoUI using textual descriptions to suppress the CRoUI recognition. 
For example, in a building CD task, 
we can select textual descriptions such as "shadows", "rivers", and "trees" as CRoUIs to inhibit the model's attention to these CRoUIs and enhance CRoI recognition.
In cases where CRoUI is not explicitly specified, empty text (" ") can be used.
We constructed text prompts for two types of text according to prompt templates and obtained text encodings $\boldsymbol{T_1}$ and $\boldsymbol{T_2}$ through the CLIP surgery's text encoder.

\paragraph{Generate Similarity Map}
Because $\boldsymbol{T_1}$ is the text encoding of CRoI and $\boldsymbol{T_2}$ is the CRoUI text encoding,
we use formulas (\ref{eq1}) and (\ref{eq2}) to calculate the similarity vector $\boldsymbol{Sim1}$ between $\boldsymbol{I_1}$ and $\boldsymbol{T_1}$
and the similarity vector $\boldsymbol{Sim2}$ between $\boldsymbol{I_2}$ and $\boldsymbol{T_2}$.\par

\begin{equation}
  \label{eq1}
  \boldsymbol{Sim1} = \mathit{Norm}[\boldsymbol{I_1}(\boldsymbol{T_1-T_2})^T] 
\end{equation}

\begin{equation}
  \label{eq2}
  \boldsymbol{Sim2} = \mathit{Norm}[\boldsymbol{I_2}(\boldsymbol{T_1-T_2})^T] 
  \end{equation}
In the above formula, \textit{Norm()} is the normalization function, and its formula is:\par
\begin{equation}
  \label{eq3}
  \mathit{Norm}(\boldsymbol{x}) = \frac{\boldsymbol{x}-\mathit{min}(\boldsymbol{x}) }{\mathit{max}(\boldsymbol{x})-\mathit{min}(\boldsymbol{x})} 
  \end{equation}

The image encoder is usually ViT\cite{vit}; thus, $\boldsymbol{I_1}$ and $\boldsymbol{I_2} \in \boldsymbol{R}^{1\times num\_token\times token\_dim}$,
$\boldsymbol{T_1}$ and $\boldsymbol{T_2} \in \boldsymbol{R}^{1\times token\_dim}$.
$\mathit{num\_token}$ is the number of tokens in ViT and can be calculated by the formula(\ref{eq4}).
In addition, $\mathit{token\_dim}$ is the dimension of each token.\par

\begin{equation}
  \label{eq4}
  \mathit{num\_token} = (\frac{\mathit{EIS}}{\mathit{PS}} )+1
\end{equation}
\par
In the above equation, \textit{EIS} is the image size of the input image encoder, and \textit{PS} is the size of each patch that divides the image into VIT during image preprocessing.
The similarity vectors $\boldsymbol{Sim1}$ and $\boldsymbol{Sim2} \in \boldsymbol{R}^{1\times token\_dim \times 1}$ are calculated to represent the similarity between each token and the CRoI.\par
Finally, the similarity vector $\boldsymbol{Sim}$ is reshaped into a two-dimensional shape to obtain a similarity map (\textit{SimM}).
However, the size of \textit{SimM}(\textit{SimMS}) obtained is $\left ( \mathit{\frac{EIS}{PS} } \times \mathit{\frac{EIS}{PS} }  \right ) $;
therefore we use the \textbf{bilinear interpolation} method to obtain the similarity map of the original image size.

\paragraph{Generate Points Prompt based on Similarity Map}
We use the threshold segmentation method to select positive/negative prompt points in the downsampled similarity map (in Fig.\ref{fig4}, the red points are positive prompt points and the blue points are negative prompt points).
The specific steps are as follows.\par
First, determine the number of positive and negative prompt points $\mathit{num\_p}$ and $\mathit{num\_n}$ according to following formula(\ref{eq5}):\par
\begin{equation}
  \label{eq5}
  \mathit{num\_p} = \mathit{min}\left \{{\mathit{sum}}(\mathit{SimM>t}),\left \lfloor \mathit{\frac{SimMS}{2} } \right \rfloor  \right \} 
  \end{equation}
In the formula(\ref{eq5}), $\mathit{t}$ is the threshold and a hyperparameter. ${\mathit{sum}}(\mathit{SimM>t})$ represents the number in the downsampled similarity map that is greater than the threshold $\mathit{t}$.
Additionally, we select the same number of negative prompt points, that is,\par
\begin{equation}
  \label{eq6}
  \mathit{num\_p} = \mathit{num\_n} 
  \end{equation}
After obtaining $\mathit{num\_p}$ and $\mathit{num\_n}$, we select the largest $\mathit{num\_p}$ points as positive prompt points and the smallest $\mathit{num\_n}$ points as negative prompt points in \textit{SimM}. 
However, the coordinates of the points selected in this way are the coordinates in \textit{SimM}, not the coordinates in the original image.
Finally, the formulas (\ref{eq7}) and (\ref{eq8}) are used to convert the coordinates of the prompt points in \textit{SimM} to the coordinates in the original image:
\begin{equation}
  \label{eq7}
  \mathit{X_0} = \mathit{\frac{H}{\sqrt{SimMS}} } \cdot X
  \end{equation}
\begin{equation}
  \label{eq8}
  \mathit{Y_0} = \mathit{\frac{W}{\sqrt{SimMS}} } \cdot Y
  \end{equation}
In the above formulas, $\mathit{X}$ and $\mathit{Y}$ are the coordinates of the prompt point in $\mathit{X_0}$ and $\mathit{Y_0}$ denotes the coordinates of the original image,
and $\mathit{H}$ and $\mathit{W}$ are the height and width of the original image, respectively.
\subsection{CRoI Segment Module}
The core component of this module is FSMs, whose main function is to receive the visual prompt of multitemporal images generated by the "Semantic Align Module" and segment the CRoI masks in the multitemporal images through the foundation segmentation model of shared weights. 
As mentioned earlier, classic FSMs, such as the segment anything model (SAM)
\cite{sam},
Although the segmentation results of FSMs represented by the SAM do not carry semantic information, it is provided by the Semantic Align Module for guidance;
therefore, when the Semantic Align Module is given a visual prompt from a CRoI, an extraction boundary about the CRoI can be generated.
Moreover, the SAM can further correct boundaries by providing some visual prompts from the CRoUI as negative samples.
Through the CRoI Segment Module, we can obtain more accurate segmentation results for the CRoI text description;
therefore, we use SAM as the FSM of this module to introduce this module in detail.\par
\begin{figure}[htbp]
  \centering
  \includegraphics[width=3.2in]{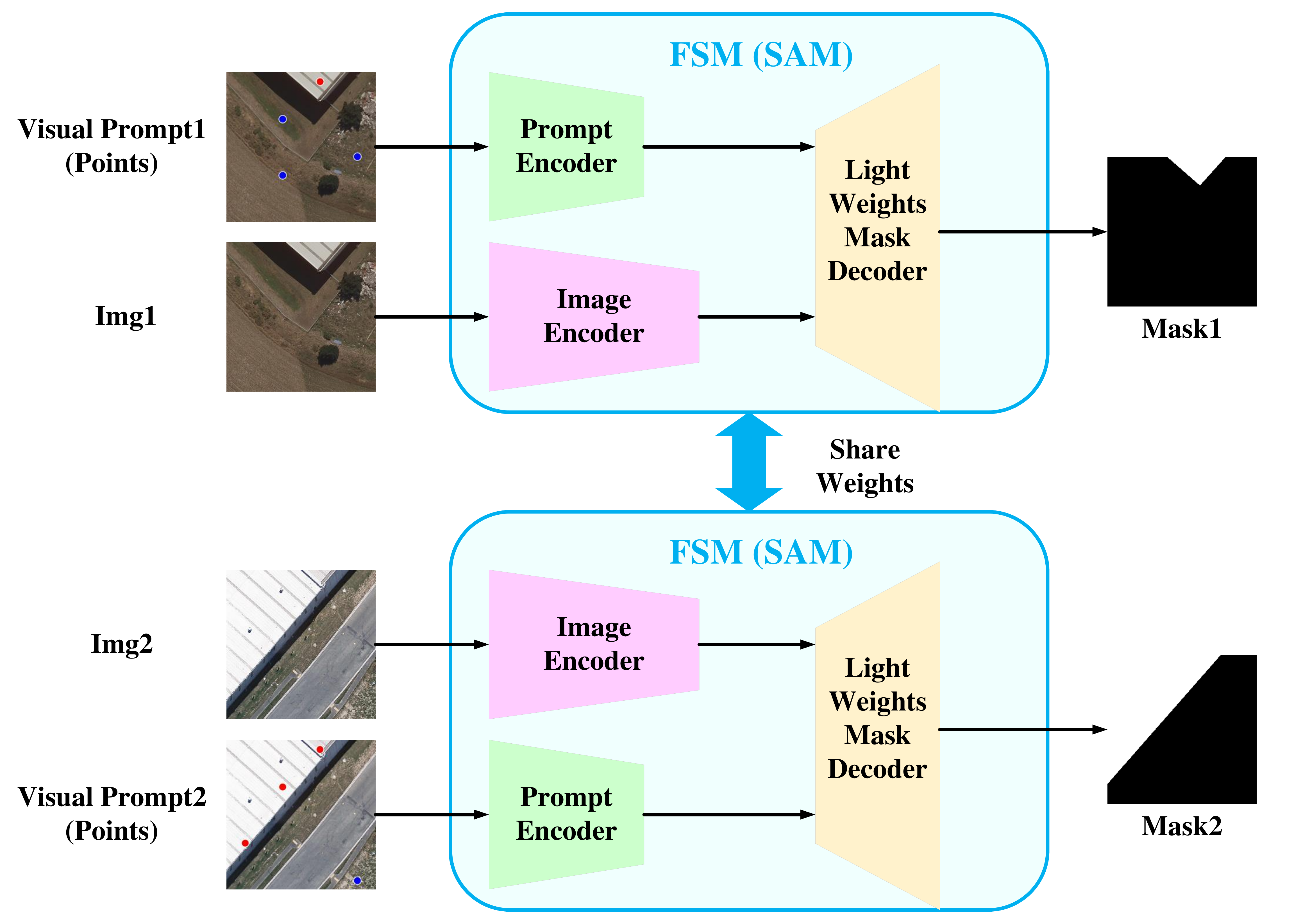}
  \caption{CRoI Segment Module (SAM as an example).}
  \label{fig5}
  \end{figure}
The SAM is an interactive segmentation framework (as shown in Fig.\ref{fig5}) that can segment the corresponding target mask based on given prompts, such as prompt points, bounding boxes or masks.	
The SAM comprises three main components: an image encoder($\phi_{i_{en}}$), a prompt encoder($\phi_{p_{en}}$), and a lightweight mask decoder($\phi_{m_{de}}$).
The SAM uses the masked autoencoder (MAE) pretraining method based on a vision transformer 
(ViT)\cite{vit} to extract image features and encode the previous prompts into embedding tokens.
Subsequently, the cross-attention mechanism in the mask decoder can promote interactions between image features and prompt embedding and ultimately generate a mask that the prompt desires.
The process is shown in Fig.\ref{fig5} and can be expressed as follows:
\begin{equation}
  \label{eq9}
  \mathit{\boldsymbol{F_{img}}} = \mathit{\phi_{i_{en}}(\boldsymbol{I}) }
\end{equation}

\begin{equation}
  \label{eq10}
  \mathit{\boldsymbol{T_{pmt}}} = \mathit{\phi_{p_{en}}(\left \{ \boldsymbol{p} \right \} ) }
\end{equation}

\begin{equation}
  \label{eq11}
  \mathit{M} = \mathit{\phi_{m_{de}}(\boldsymbol{F_{img}}+\boldsymbol{F_{c\_mask}},\boldsymbol{[T_{out},T_{pmt}]} ) }
\end{equation}

where $\boldsymbol{I} \in \boldsymbol{R}^{H\times W \times 3}$ represents the original image,
$\boldsymbol{F_{img}} \in \boldsymbol{R}^{h\times w \times c}$ represents the image features encoded by the SAM image encoder,
and $\boldsymbol{{p}}$ contains sparse prompts and can contain positive/negative sample points(also in Fig.\ref{fig5}, the red points are positive prompt points and the blue points are negative prompt points), bounding boxes or masks.
$\boldsymbol{T_{pmt}} \in \boldsymbol{R}^{k\times c}$ represents the sparse prompt token encoded by the SAM prompt encoder.
In addition, $\boldsymbol{F_{c\_mask}} \in \boldsymbol{R}^{h\times w \times c}$ is the representation of the coarse mask, which is an optional input of the SAM.
$\boldsymbol{T_{out}} \in \boldsymbol{R}^{5\times c}$ consists of preinserted learnable tokens representing four different masks and their corresponding IoU predictions. 
Finally, $\mathit{M}$ is the corresponding prediction mask. 
With respect to the AUWCD, only one output is needed; thus, we choose the one with the highest prediction score as the final prediction result.\par

\subsection{CD(CD) Module}
The main function of this module is to perform the change process on the CRoI mask in the multitemporal image and finally obtain the CRoI change map. 
We assume that we obtain the mask of multiple temporal images CRoI $\mathit{\left \{ M_{t1}, M_{t2},\cdots, M_{tn}\right \} }$ in a certain area through the foundation segment module. 
The change map obtained by the change process for any two phases can be expressed by the formula(\ref{eq12}):\par
\begin{equation}
  \label{eq12}
  \mathit{ChangeMap_{(t_i,t_j)}}=(M_{t_i}\cup M_{t_j} ) \setminus (M_{t_i}\cap  M_{t_j} )
\end{equation}

In the formula, $\mathit{ChangeMap_{(t_i,t_j)}}$ represents the change map of the image at time $\mathit{t_j}$ in some area relative to the image at time $\mathit{t_i}$ in the same area.
$\setminus$ is the difference set symbol, and $A \setminus B$ represents the difference set of A to B.\par
Notably, the VLM and VSM include CLIP surgery and SAM. We can also choose different change process methods to obtain a higher-quality change map.\par

\section{Experiments}
Experimental validations were conducted using the instantiated framework described above to assess the effectiveness of the AUWCD framework.
We first outline the experimental setup, including the dataset, evaluation metrics, and implementation details, followed by presenting the experimental results.  
Finally, a brief summary and analysis of the experimental outcomes are provided.\par
\subsection{Databases}
\subsubsection{BCDD Dataset}
\begin{figure}[htbp]
  \centering
  \includegraphics[width=3.2in]{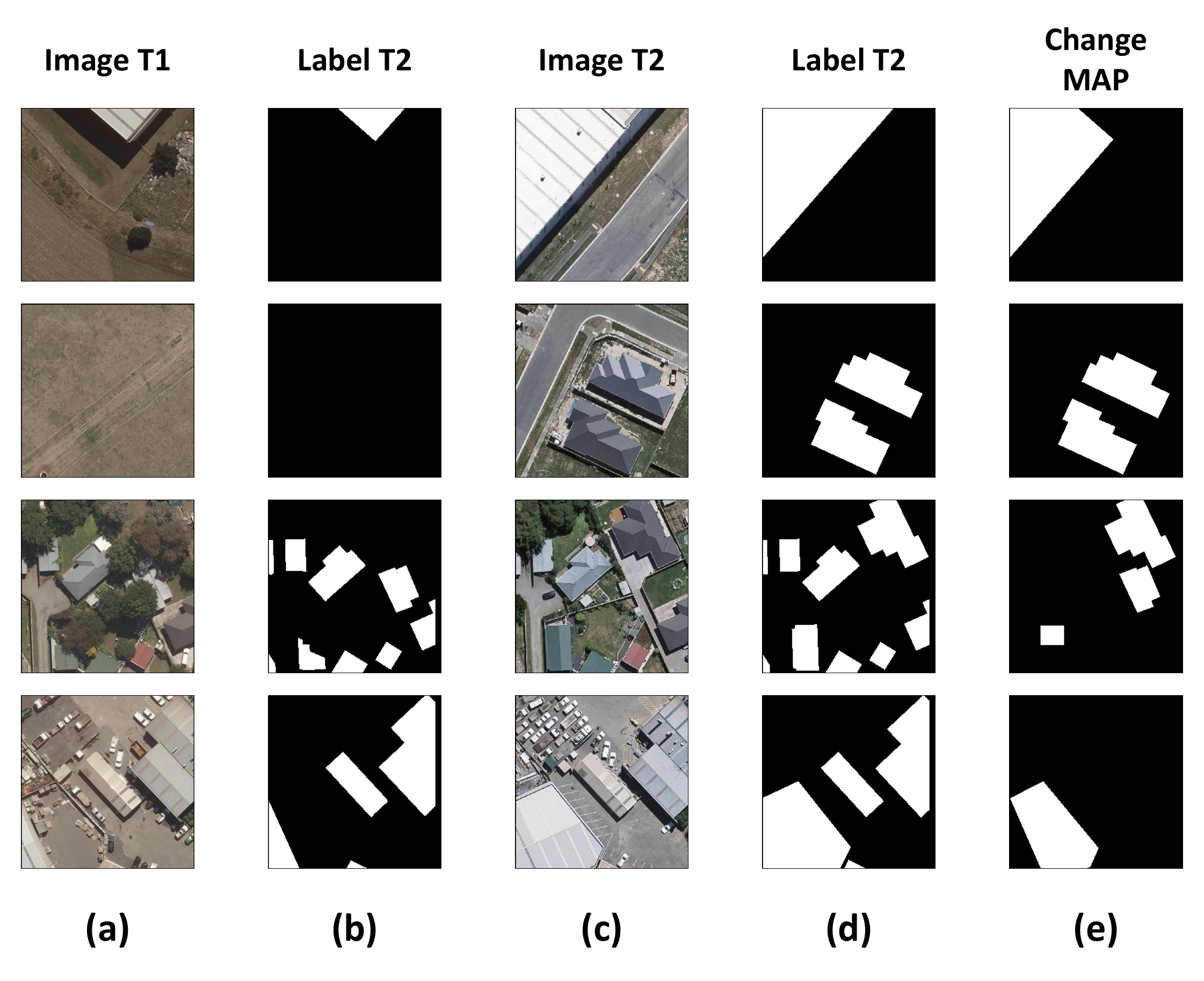}
  \caption{Multitemporal images that were selected from the BCDD dataset.}
  \label{fig6}
  \end{figure}

The BCDD dataset\cite{bcdd} (as shown in Fig.\ref{fig6}) encompasses the region affected by a 6.3 magnitude earthquake in February 2011 in Christchurch, New Zealand.
This dataset comprises two image scenes captured at the same location in 2012 and 2016, accompanied by semantic labels and CD labels for buildings. 
To manage the large image size of 32,507 $\times$ 15,354 pixels, we divided each image into non-overlapping 256 $\times$ 256 pixels image pairs, resulting in a total of 7,434 pairs.\par
\subsubsection{Second Dataset}
\begin{figure}[htbp]
  \centering
  \includegraphics[width=3.2in]{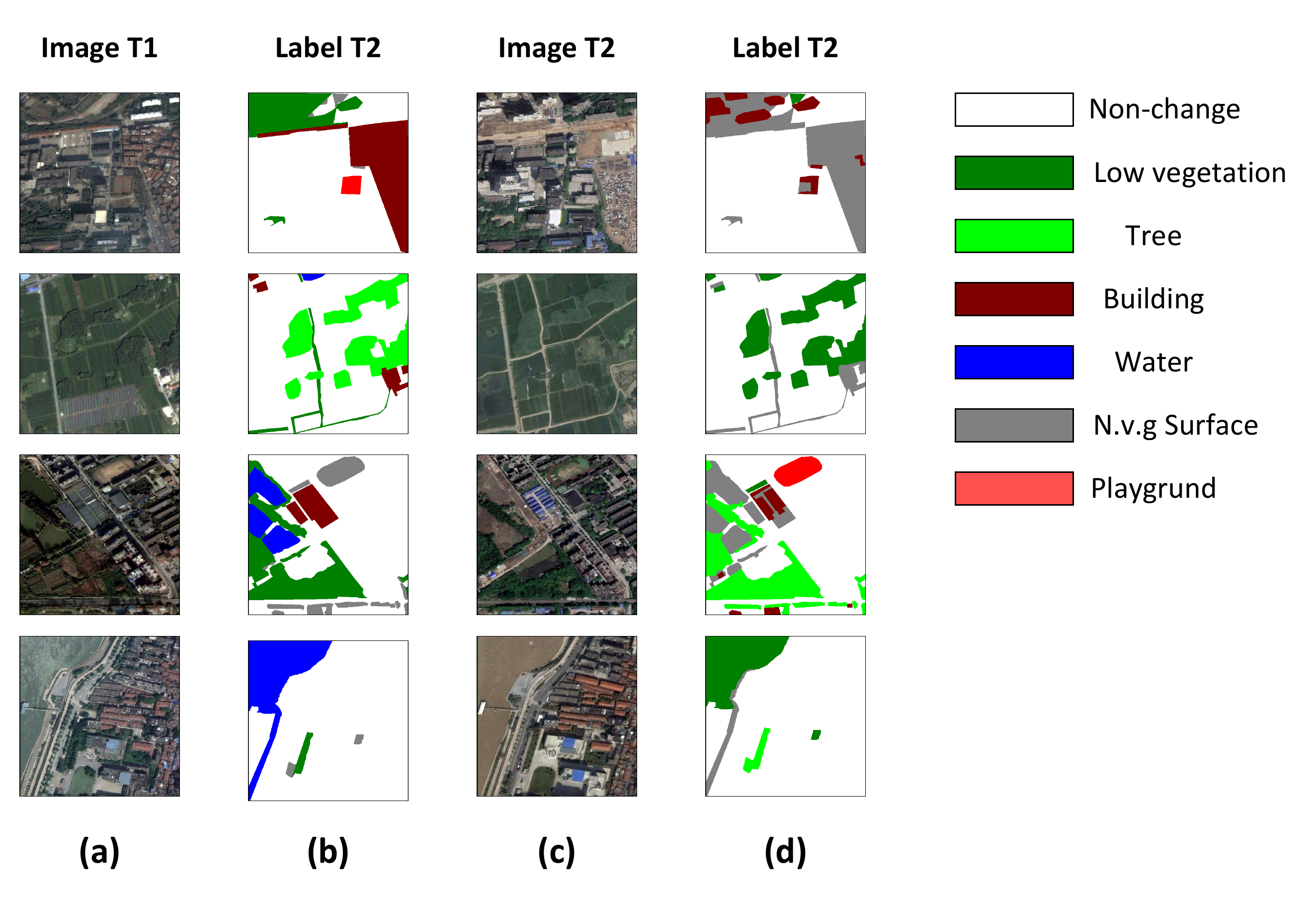}
  \caption{Multitemporal images that were selected from the SECOND dataset.}
  \label{fig7}
  \end{figure}

The SECOND dataset\cite{second} is a semantic CD dataset (as illustrated in Fig.\ref{fig7}).
It consists of 4,662 pairs of aerial images collected from various platforms and sensors covering cities such as Hangzhou, Chengdu, and Shanghai.
Each image in the dataset has a size of 512 $\times$ 512 pixels and is annotated at the pixel level.
The dataset focuses on six land cover categories, namely, \textbf{non-vegetation}, \textbf{trees}, \textbf{low vegetation}, \textbf{water}, \textbf{buildings}, and \textbf{playgrounds}.
We established a baseline CRoI based on the land cover categories in the first temporal phase.
Using these six CRoIs, the dataset was divided into six CRoI categories. 
Subsequently, the images were utilized to create the training set, validation set, and test set in an 8:1:1 ratio.\par
\subsection{Metrics}
To quantitatively validate the effectiveness of the proposed method, we employed four common accuracy metrics commonly used in CD tasks\cite{dasnet}:
\textbf{precision}, \textbf{recall}, \textbf{F1 score (F1)}, and \textbf{overall accuracy (OA)}. The formulas for these metrics are as follows:\par

\begin{equation}
  \label{eq13}
  \mathit{Precision}=\mathit{\frac{TP}{TP+FP} }  
\end{equation}

\begin{equation}
  \label{eq14}
  \mathit{Recall}=\mathit{\frac{TP}{TP+FN} }  
\end{equation}

\begin{equation}
  \label{eq15}
  \mathit{F1}=\mathit{\frac{2 \cdot TP \cdot Recall}{Precision+Recall} }  
\end{equation}

\begin{equation}
  \label{eq16}
  \mathit{OA}=\mathit{\frac{TP+TN}{TP+TN+FP+FN} }  
\end{equation}
where TP is the number of true positives, FP is the number of false positives, TN is the number of true negatives, and FN is the number of false negatives.\par
In addition, this paper utilizes the \textbf{point accuracy (PAcc)} metric to quantify the effectiveness of the point prompt generated by the "Semantic Align Module".
The computation formula for the PAcc is as follows:\par
\begin{equation}
  \label{eq17}
  \mathit{PAcc}=\mathit{\frac{P_c+N_c}{P+N} }  
\end{equation}
where $\mathit{P}$ is the total number of positive points generated by CLIP surgery and $\mathit{N}$ is the total number of negative points generated by CLIP surgery,
$\mathit{P_c}$ is the number of positive points whose corresponding label is 1 (CRoI),
$\mathit{N_c}$ is the number of negative points whose corresponding label is 0 (CRoUI).
\subsection{Comparative Experiment}
The purpose of this comparative experiment was to quantitatively compare the effects of the AUWCD framework and advanced methods in the ViFi-CD paradigm on different CRoI CD tasks
without major modifications to the model or method, the SeFi-CD paradigm can better adapt to complex and changeable CRoI CD than the ViFi-CD paradigm.
The experimental details and results are as follows.\par
\subsubsection{Experimental Design Ideas}
The idea for the design of this experiment is that in a multicategory CD dataset, various categories represent CRoIs for different tasks.
We selected several major supervised baselines dominated by supervised learning in the ViFi-CD paradigm as comparison objects.\par

For these supervised baselines, we provide training samples for only a certain category in the dataset and do not provide training samples for the other categories. 
Thus, the baseline obtained after training can be regarded as a model dedicated to a specific CRoI CD task and other categories can be regarded as different CD tasks focused on CRoIs. 

Finally, we use the trained baselines to perform CD tasks on all CRoIs categories and use the above indicators to quantify the detection effect; for AUWCD, we input only the text of each category in the dataset into the model.
To prompt the model to complete different categories of CD tasks, the above indicators are used to quantify the effect of CD;
finally, the effects of the AUWCD and these supervised baselines are compared by comparing indicator values.\par

\subsubsection{Implementation Details}
For multicategory CD datasets, we choose the SECOND dataset, which contains six classes for semantic CD,
as the experimental dataset to verify the effect of the AUWCD framework.
We compare the performance of the AUWCD with that of advanced CD methods on the ViFi-CD dataset. \par

For the methods in ViFi-CD, we chose several advanced supervised baselines from the past five years as objects of comparison:
FC-EF\cite{fully_convolutional_siamese_networks_for_change_detection}, 
FC-Siam-conc\cite{fully_convolutional_siamese_networks_for_change_detection},
FC-Siam-diff\cite{fully_convolutional_siamese_networks_for_change_detection}, 
DASNet\cite{dasnet}, 
BIT\cite{bit}, 
SwinSUNet\cite{swinsunet}, 
ICIFNet\cite{icifnet}, 
DMINet\cite{dminet},
ELGCNet\cite{elgcnet},
ChangeBind\cite{changebind}.

All supervised baselines were implemented using the PyTorch backend and were supported by NVIDIA RTX A6000 GPUs.
We used the Adam optimizer with an initial learning rate of 0.00001 for all the models and set the batch size to 8. \par

For the AUWCD framework, we needed to change only the CRoI text prompt to directly test each of the six CRoIs separately.
The various hyperparameters of the point prompt generation process in CLIP surgery used the best hyperparameters obtained from subsequent ablation experiments and used the largest and best-performing model, CSVIT-L-14. The FSM module also utilizes the sam\_vit\_h model.
Both CSVIT-L-14 and sam\_vit\_h use pretrained model parameters.
The text prompts for each CRoI category were set as follows:
"low vegetation land", "tree", "building roofs", "waters", "non-vegetation land" and "football field".
These datasets corresponded to the original low vegetation, tree, building, water, n.v.g. surface, and playground categories, respectively. \par

To eliminate chance, we conducted three independent sets of comparative experiments, and the CRoI categories that provided training samples were divided into "tree", "water", and "playground".
Moreover, to demonstrate our effects quantitatively, we chose the F1 score, which is the most important indicator for CD tasks, as a quantitative indicator.\par
\subsubsection{Result}

\paragraph{"Tree" was used as the training category to provide training samples}
The experimental results for the other categories that did not provide training samples are shown in Table \ref{tab:table2}.\par
\begin{table*}
  \caption{Experimental results of tree as a training category that provides training samples and other categories that do not provide training samples. The indicators in the table are F1 scores(\%). \label{tab:table2}}
  \centering
  \begin{tblr}{
    row{even} = {c},
    row{3} = {c},
    row{5} = {c},
    row{7} = {c},
    row{9} = {c},
    row{11} = {c},
    cell{1}{2} = {c},
    cell{1}{3} = {c},
    cell{1}{4} = {c},
    cell{1}{5} = {c},
    cell{1}{6} = {c},
    cell{1}{7} = {c},
    cell{1}{8} = {c},
    hline{1-2,12-13} = {-}{},
  }
  \diagbox{\textbf{Baseline}}{\textbf{CRoI}} & {\textbf{Tree}\\\textbf{(Sup.)}} & \textbf{Low Veg.} & \textbf{Building} & \textbf{Waters} & \textbf{N.V.g. Surface}& \textbf{Playground} \\
  \textbf{FC-EF}(2018)                   & 2.38                             & 9.02              & 21.39             & 0.29            & 19.50                               & 0.24                \\
  \textbf{FC-Siam-conc}(2018)            & 0.37                             & 0.54              & 0.70              & 0.12            & 0.41                                & 0.32                \\
  \textbf{FC-Siam-diff}(2018)            & 4.44                             & 13.20             & 27.01             & 0.55            & 22.76                               & 0.24                \\
  \textbf{DASNet}(2020)                  & \textbf{37.97}                   & 16.56             & 4.65              & 1.22            & 14.37                               & 0.47                \\
  \textbf{BIT}(2021)                     & 16.83                            & 4.66              & 0.68              & 0.42            & 3.43                                & 1.03                \\
  \textbf{SwinSUnet}(2022)               & 0.90                             & 2.18              & 0.77              & 1.16            & 1.20                                & 0.00                \\
  \textbf{ICIFNet}(2022)                 & 19.82                            & 4.29              & 1.01              & 0.40            & 3.64                                & 0.67                \\
  \textbf{DMINet}(2022)                  & 16.90                            & 3.89              & 0.78              & 0.25            & 3.49                                & 1.13                \\
  \textbf{ELGCNet}(2024)                 & 0.49                             & 0.44              & 0.09              & 0.86            & 0.37                                & 0.00               \\         
  \textbf{ChangeBind}(2024)              & 14.91                            & 7.13              & 3.18              & 5.65            & 5.65                                & 0.90               \\
  \textbf{Ours}                          & 8.21                             & \textbf{23.43}    & \textbf{34.17}    & \textbf{2.21}   & \textbf{27.80}                      & \textbf{1.19}       
  \end{tblr}
  \end{table*}

\paragraph{"Water" was used as the training category to provide training samples}
The experimental results for the other categories that did not provide training samples are shown in Table \ref{tab:table3}.\par
\begin{table*}
  \caption{Experimental results of water as a training category that provides training samples and other categories that do not provide training samples. The indicators in the table are F1 scores(\%).\label{tab:table3}}
  \centering
  \begin{tblr}{
    row{even} = {c},
    row{3} = {c},
    row{5} = {c},
    row{7} = {c},
    row{9} = {c},
    row{11} = {c},
    cell{1}{2} = {c},
    cell{1}{3} = {c},
    cell{1}{4} = {c},
    cell{1}{5} = {c},
    cell{1}{6} = {c},
    cell{1}{7} = {c},
    cell{1}{8} = {c},
    hline{1-2,12-13} = {-}{},
  }
  \diagbox{\textbf{Baseline}}{\textbf{CRoI}} & {\textbf{\textbf{Waters}}\\\textbf{(Sup.)}} & \textbf{Low Veg.} & \textbf{Tree} & \textbf{Building} & \textbf{N.V.g. Surface} & \textbf{Playground} \\
  \textbf{FC-EF}(2018)                   & 0.45                                        & 9.04              & 3.64          & 13.41             & 15.74                   & 0.28                \\
  \textbf{FC-Siam-conc}(2018)            & 0.74                                        & 16.07             & 5.42          & 21.00             & 24.77                   & 0.37                \\
  \textbf{FC-Siam-diff}(2018)            & 0.00                                        & 0.17              & 0.05          & 0.17              & 0.23                    & 0.00                \\
  \textbf{DASNet}(2020)                  & \textbf{16.48}                              & 3.56              & 0.56          & 1.24              & 1.36                    & 0.45                \\
  \textbf{BIT}(2021)                     & 1.43                                        & 0.07              & 0.01          & 0.00              & 0.03                    & 0.00                \\
  \textbf{SwinSUNet}(2022)               & 0.00                                        & 0.00              & 0.00          & 0.00              & 0.00                    & 0.00                \\
  \textbf{ICIFNet}(2022)                 & 3.99                                        & 0.10              & 0.02          & 0.02              & 0.11                    & 0.00                \\
  \textbf{DMINet}(2022)                  & 5.50                                        & 0.29              & 0.02          & 0.08              & 0.14                    & 0.00                \\
  \textbf{ELGCNet}(2024)                 & 0.00                                        & 0.00              & 0.00          & 0.00              & 0.00                    & 0.00            \\            
  \textbf{ChangeBind}(2024)              & 2.09                                        & 0.34              & 0.23          & 0.40              & 0.32                    & 0.00               \\
  \textbf{Ours}                          & 2.21                                        & \textbf{23.43}    & \textbf{8.21} & \textbf{34.17}    & \textbf{27.80}          & \textbf{1.19}       
  \end{tblr}
  \end{table*}

\paragraph{"Playground" was used as the training category to provide training samples}
The experimental results of other categories that did not provide training samples are shown in Table \ref{tab:table4}.\par
\begin{table*}
  \caption{Experimental results of playground as a training category that provides training samples and other categories that do not provide training samples. The indicators in the table are F1 scores(\%).\label{tab:table4}}
  \centering
  \begin{tblr}{
    row{even} = {c},
    row{3} = {c},
    row{5} = {c},
    row{7} = {c},
    row{9} = {c},
    row{11} = {c},
    cell{1}{2} = {c},
    cell{1}{3} = {c},
    cell{1}{4} = {c},
    cell{1}{5} = {c},
    cell{1}{6} = {c},
    cell{1}{7} = {c},
    cell{1}{8} = {c},
    hline{1-2,12-13} = {-}{},
  }
  \diagbox{\textbf{Baseline}}{\textbf{CRoI}} & {\textbf{\textbf{Playground}}\\\textbf{(Sup.)}} & \textbf{Low Veg.} & \textbf{Tree} & \textbf{Building} & \textbf{Waters} & \textbf{\textbf{N.V.g.~Surface}} \\
  \textbf{FC-EF}(2018)                   & 0.23                                            & 8.64              & 1.95          & 21.86             & 0.21            & 19.33                            \\
  \textbf{FC-Siam-conc}(2018)            & 0.16                                            & 0.71              & 1.28          & 0.73              & 0.39            & 0.75                             \\
  \textbf{FC-Siam-diff}(2018)            & 0.25                                            & 12.99             & 3.90          & 27.64             & 0.45            & 22.74                            \\
  \textbf{DASNet}(2020)                  & 22.16                                           & 1.14              & 0.09          & 0.23              & 0.46            & 0.77                             \\
  \textbf{BIT}(2021)                     & 14.10                                           & 0.40              & 0.01          & 0.02              & 0.11            & 0.30                             \\
  \textbf{SwinSUNet}(2022)               & 0.92                                            & 0.00              & 0.00          & 0.00              & 0.00            & 0.00                             \\
  \textbf{ICIFNet}(2022)                 & 4.02                                            & 0.27              & 0.02          & 0.02              & 0.00            & 0.17                             \\
  \textbf{DMINet}(2022)                  & \textbf{36.87}                                  & 0.53              & 0.00          & 0.30              & 0.00            & 0.73                             \\
  \textbf{ELGCNet}(2024)                 & 0.00                                            & 0.00              & 0.00          & 0.00              & 0.00            & 0.00                    \\   
  \textbf{ChangeBind}(2024)              & 0.33                                            & 0.01              & 0.00           & 0.02             & 0.00            & 0.02               \\
  \textbf{Ours}                          & 1.19                                            & \textbf{23.43}    & \textbf{8.21} & \textbf{34.17}    & \textbf{2.21}   & \textbf{27.80}                   
  \end{tblr}
  \end{table*}
  \par

The experimental results show that the performance of the AUWCD can exceed that of several CD supervision baselines compared with the CRoI categories that provide training samples for each supervision baseline;
however, a certain gap remains compared with the current, more advanced supervision baselines.
Compared with the performance that is not provided for the training sample categories, the performance of the AUWCD exceeds that of these supervised baselines by simply modifying the CRoI prompt text without additional training.
The F1 score exceeds these advanced supervised baselines by an average of 5.01 percentage points, and the highest exceeds it by 13.17 percentage points.\par
\subsection{Visualization Experiment}
A visual experiment of the AUWCD is performed in this section.
The main purpose is to visually demonstrate the process and effect of CD via the AUWCD. The experimental details and results are as follows: \par
\subsubsection{Implementation Details}

\paragraph{Visualization effect of the AUWCD on a single CRoI CD task (GT is used for comparison, but the category is single)}
We display the visual results of building CD on the BCDD dataset to show the intermediate process and effects of AUWCD CD and the final CD results.
Consistent with the parameter settings obtained after the subsequent ablation experiment, VLM selected CSVIT-L-14, the threshold t was set to 0.65, and the CRoI prompt word was set to "iron building house roofs". For FSM, we also selected sam\_vit\_h.
Both CSVIT-L-14 and sam\_vit\_h also used pretrained model parameters. \par

\paragraph{Visualization effect of the AUWCD on multiple CRoI CD tasks (no information for comparison but rather multiple categories)}
In the BCDD dataset, we demonstrate the CD effect of different CRoIs on the same pair of dual-temporal images.
In the experiment, we selected several ground objects for display.
The selected ground objects and corresponding prompt words are shown in Table \ref{tab:table5}:\par
\begin{table}[htbp]
  \caption{An Example of a Table\label{tab:table5}}
  \centering
  \begin{tblr}{
    cells = {c},
    hline{1-2,7} = {-}{},
  }
  \textbf{CRoI Name} & \textbf{Prompt Text} \\
  Building           & "house roofs"        \\
  River              & "river"              \\
  Tree               & "trees"              \\
  Car                & "cars"               \\
  Road               & "asphalt road"       
  \end{tblr}
  \end{table}

\subsubsection{Results}
\paragraph{The visualization effect of the AUWCD on a single CRoI CD task}
\begin{figure*}[htbp]
  \centering
  \includegraphics[width=7in]{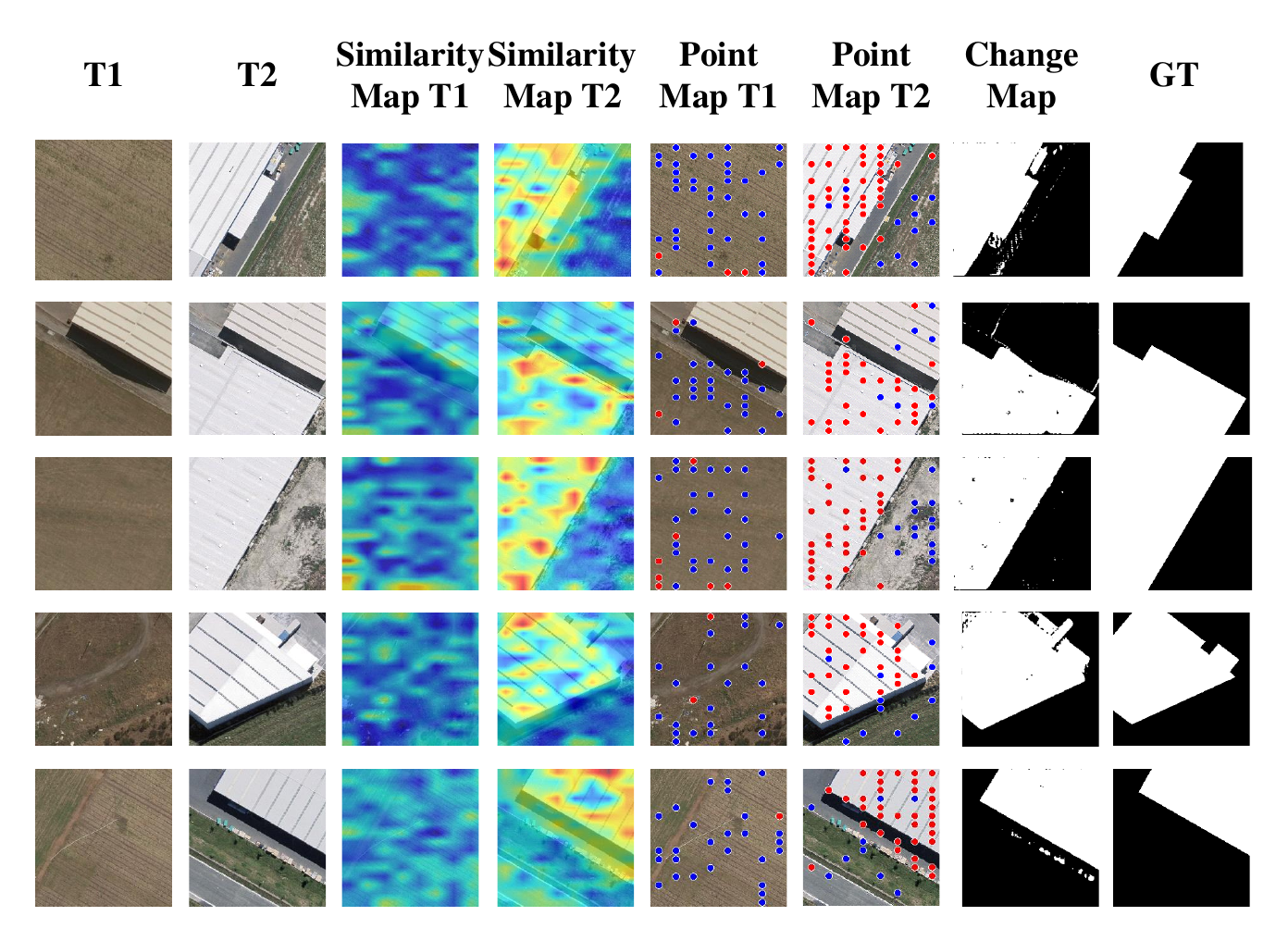}
  \caption{Visualization results of single CRoI (building) CD in the BCDD dataset.}
  \label{fig8*}
\end{figure*}

The experimental results(Fig.\ref{fig8*}) show the Similarity Map and Point Map generated in the Semantic Align Module to reflect the intermediate process of CD by AUWCD and its effect;
and we show the CD effect of the AUWCD by showing the final CD results.\par

Fig.\ref{fig8*} shows that under the given text prompt "iron building house roofs",
from the Similarity Map heatmap is higher in the building area, which indicates that the model can pay attention to the CRoIs we are interested in.
Points also accurately fall in areas outside the building;
however, there are still some points that fail to fall in the correct area and optimizing these points is a future research direction.
Finally, from the comparison between the generated change map and GT labels, although a gap exists between the generated change map and GT labels, the overall effect is still acceptable.

\paragraph{The visualization effect of the AUWCD for multiple CRoI CD tasks}

\begin{table*}[htbp]
  \caption{Visualization results of various CRoI CD in the BCDD dataset\label{tab:table6}}
  \centering
  \begin{tblr}{cells = {c},cell{2}{1} = {r=3}{},cell{5}{1} = {r=4}{},hline{1-2,5,9} = {-}{}}
  \textbf{Image}                                        & \textbf{CRoI}                  & {\textbf{Similarity}\\\textbf{Map T1}} & {\textbf{Similarity}\\\textbf{Map T2}} & {\textbf{Point}\\\textbf{Map T1}} & {\textbf{Point}\\\textbf{Map T2}} & {\textbf{Change}\\\textbf{Map}} \\
  {\begin{minipage}{.1\textwidth} \includegraphics[width=0.75in]{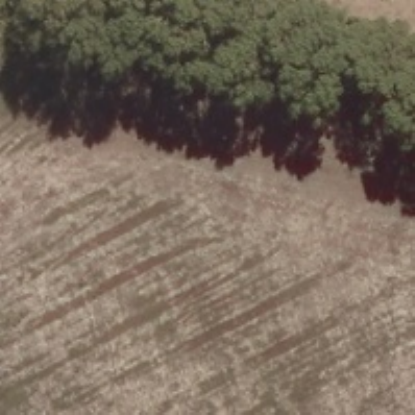} \end{minipage}\\
  ~~\\\textbf{T1} \\  ~~\\
  \begin{minipage}{.1\textwidth} \includegraphics[width=0.75in]{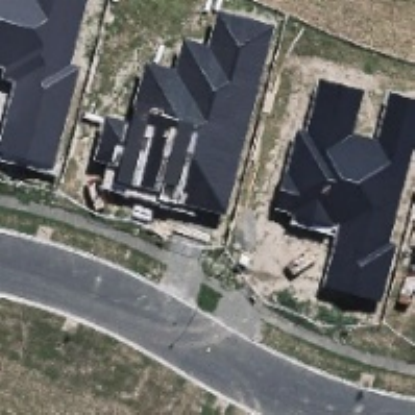} \end{minipage}\\
  ~~\\\textbf{T2}}              & \textbf{\textbf{Road}} 
                                                                                        & \begin{minipage}{.1\textwidth} \includegraphics[width=0.75in]{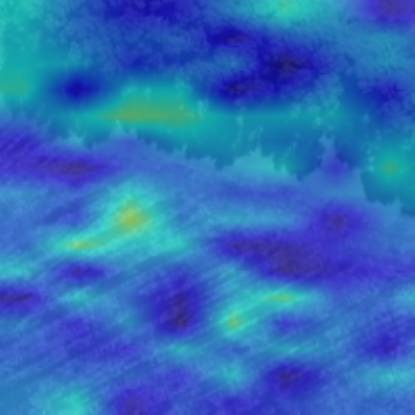} \end{minipage} 
                                                                                        & \begin{minipage}{.1\textwidth} \includegraphics[width=0.75in]{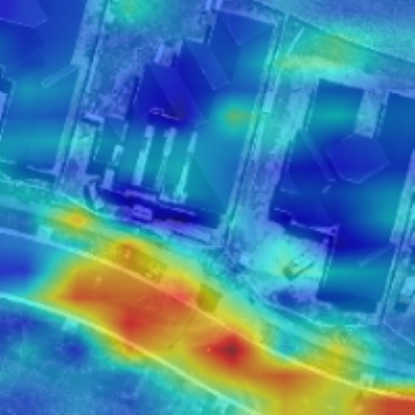} \end{minipage} 
                                                                                        & \begin{minipage}{.1\textwidth} \includegraphics[width=0.75in]{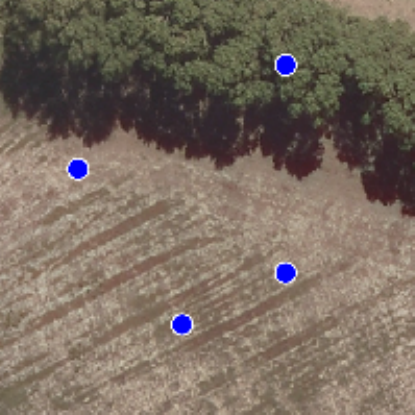} \end{minipage}  
                                                                                        & \begin{minipage}{.1\textwidth} \includegraphics[width=0.75in]{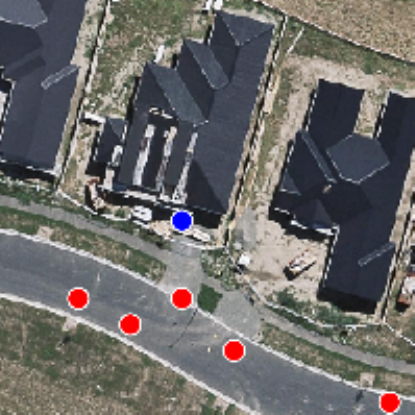} \end{minipage} 
                                                                                        & \begin{minipage}{.1\textwidth} \includegraphics[width=0.75in]{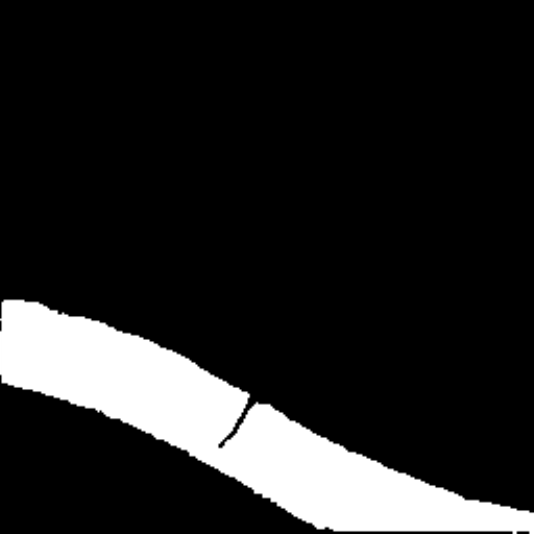} \end{minipage}                               \\
                                                        & \textbf{\textbf{Building}}     
                                                                                        & \begin{minipage}{.1\textwidth} \includegraphics[width=0.75in]{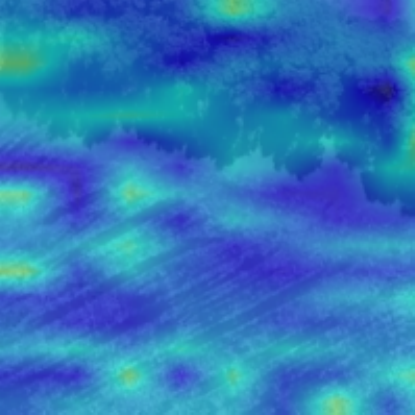} \end{minipage}
                                                                                        & \begin{minipage}{.1\textwidth} \includegraphics[width=0.75in]{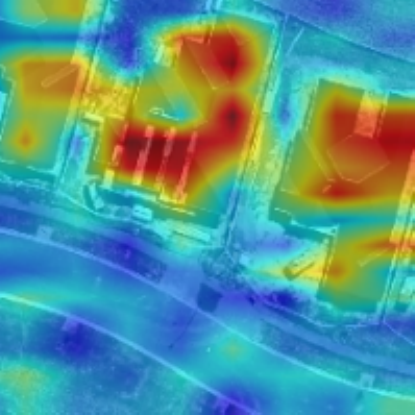} \end{minipage}                                     
                                                                                        & \begin{minipage}{.1\textwidth} \includegraphics[width=0.75in]{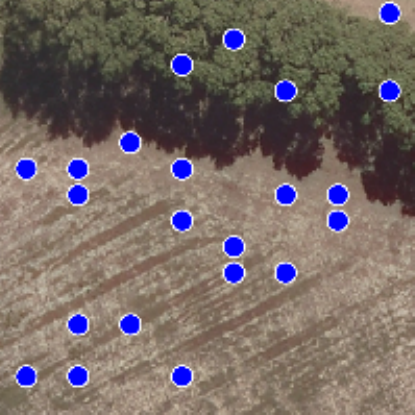} \end{minipage}                                
                                                                                        & \begin{minipage}{.1\textwidth} \includegraphics[width=0.75in]{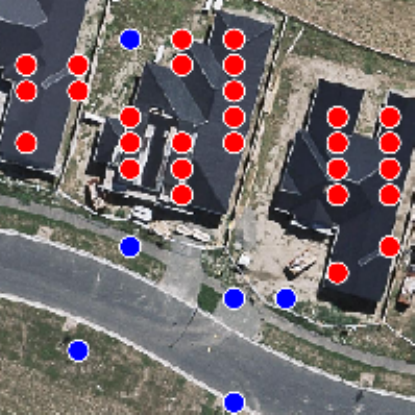} \end{minipage}                                
                                                                                        & \begin{minipage}{.1\textwidth} \includegraphics[width=0.75in]{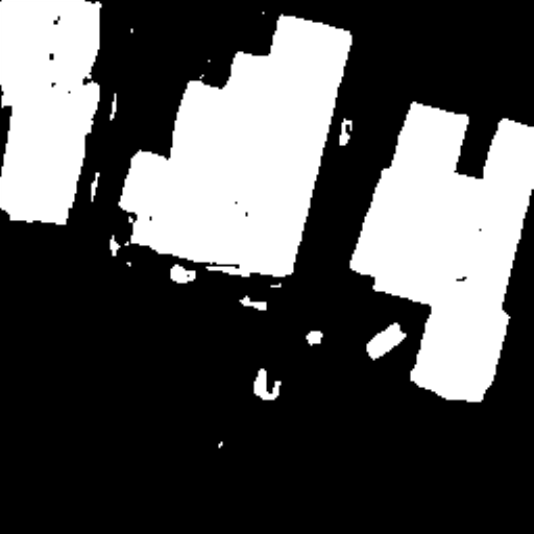} \end{minipage}                             \\
                                                        & \textbf{\textbf{Trees}}       
                                                                                        & \begin{minipage}{.1\textwidth} \includegraphics[width=0.75in]{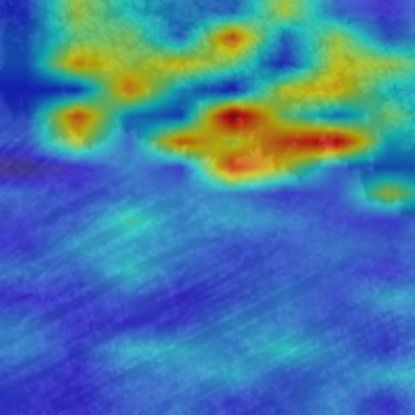} \end{minipage}                                  
                                                                                        & \begin{minipage}{.1\textwidth} \includegraphics[width=0.75in]{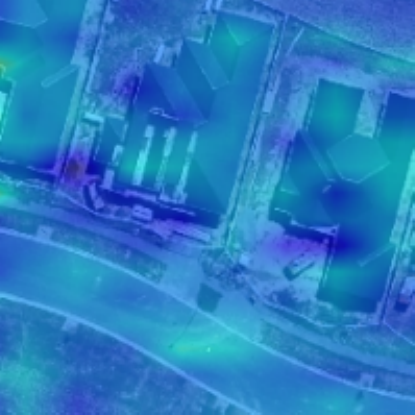} \end{minipage}                                    
                                                                                        & \begin{minipage}{.1\textwidth} \includegraphics[width=0.75in]{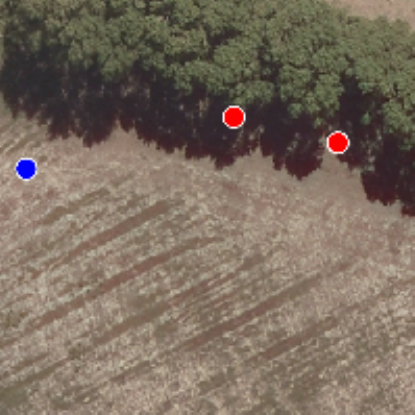} \end{minipage}                               
                                                                                        & \begin{minipage}{.1\textwidth} \includegraphics[width=0.75in]{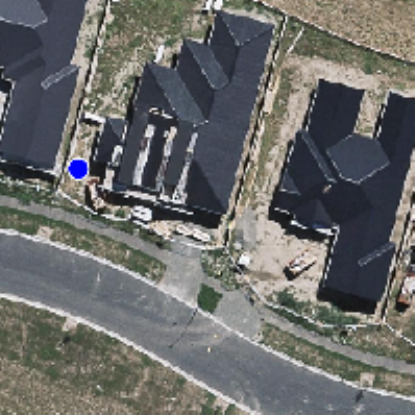} \end{minipage}                               
                                                                                        & \begin{minipage}{.1\textwidth} \includegraphics[width=0.75in]{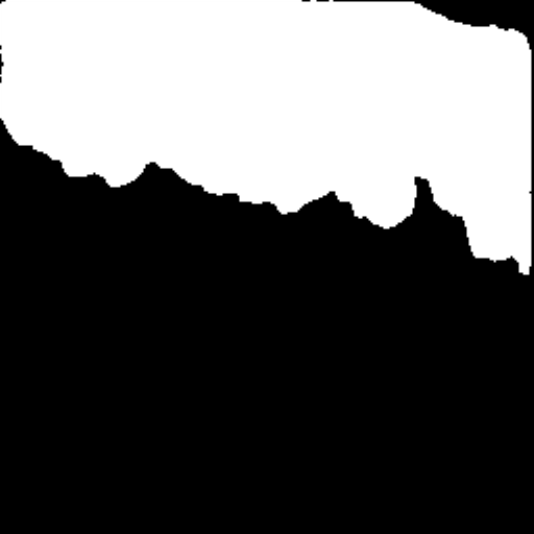} \end{minipage}                             \\
  {\begin{minipage}{.1\textwidth} \includegraphics[width=0.75in]{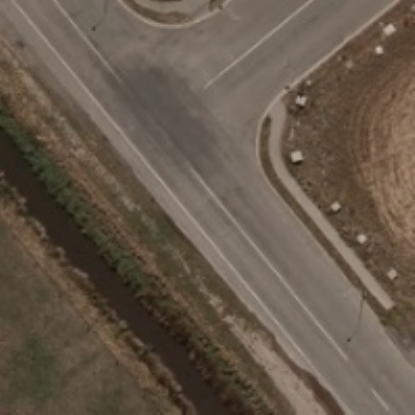} \end{minipage}\\
  ~~\\ \textbf{T1} \\ ~~\\
  \begin{minipage}{.1\textwidth} \includegraphics[width=0.75in]{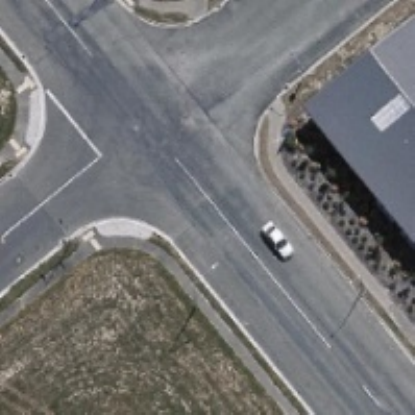} \end{minipage} \\
  ~~\\ \textbf{T2}} 
                                                      & \textbf{\textbf{Cars}}         
                                                                                        & \begin{minipage}{.1\textwidth} \includegraphics[width=0.75in]{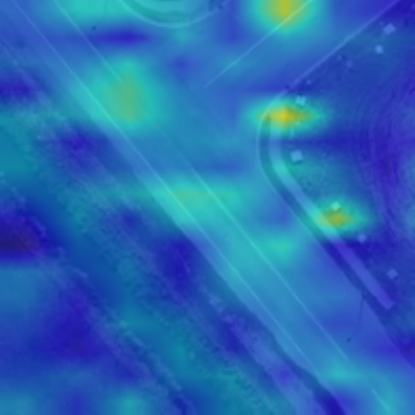} \end{minipage}                                    
                                                                                        & \begin{minipage}{.1\textwidth} \includegraphics[width=0.75in]{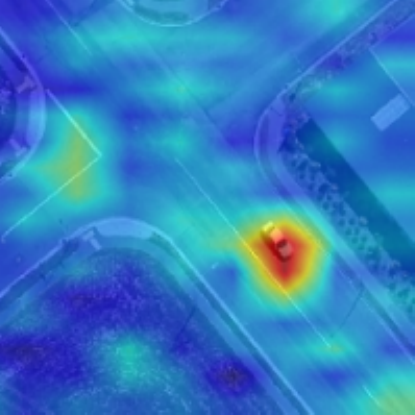} \end{minipage}                                    
                                                                                        & \begin{minipage}{.1\textwidth} \includegraphics[width=0.75in]{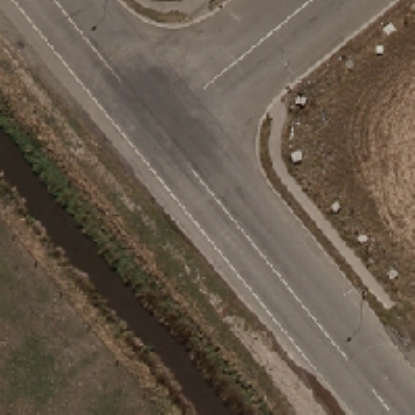} \end{minipage}                               
                                                                                        & \begin{minipage}{.1\textwidth} \includegraphics[width=0.75in]{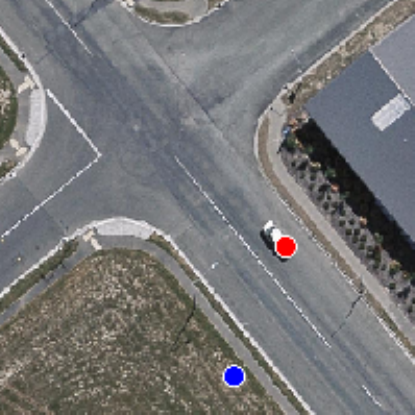} \end{minipage}                               
                                                                                        & \begin{minipage}{.1\textwidth} \includegraphics[width=0.75in]{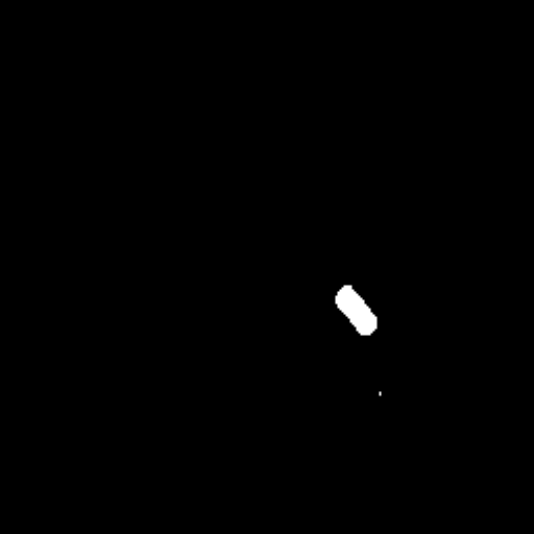} \end{minipage}                             \\
                                                        & \textbf{\textbf{Building}}     
                                                                                        & \begin{minipage}{.1\textwidth} \includegraphics[width=0.75in]{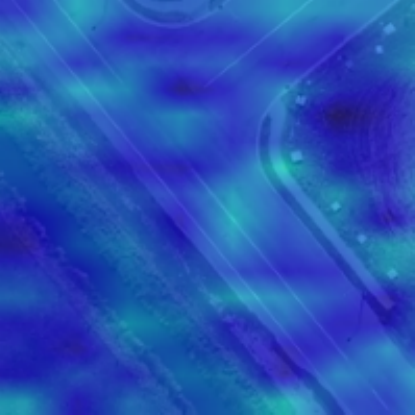} \end{minipage}                                    
                                                                                        & \begin{minipage}{.1\textwidth} \includegraphics[width=0.75in]{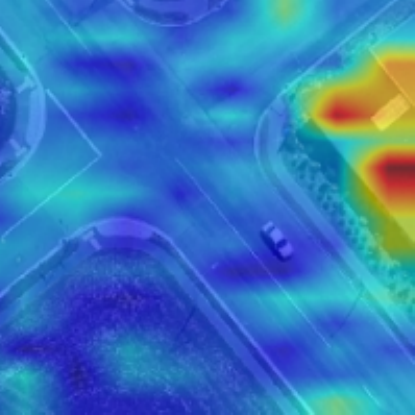} \end{minipage}                                    
                                                                                        & \begin{minipage}{.1\textwidth} \includegraphics[width=0.75in]{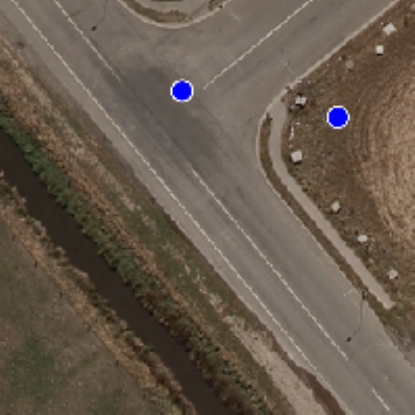} \end{minipage}                               
                                                                                        & \begin{minipage}{.1\textwidth} \includegraphics[width=0.75in]{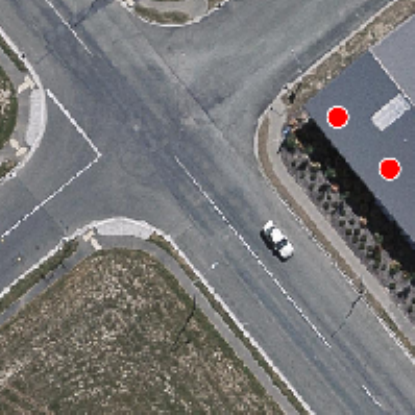} \end{minipage}                               
                                                                                        & \begin{minipage}{.1\textwidth} \includegraphics[width=0.75in]{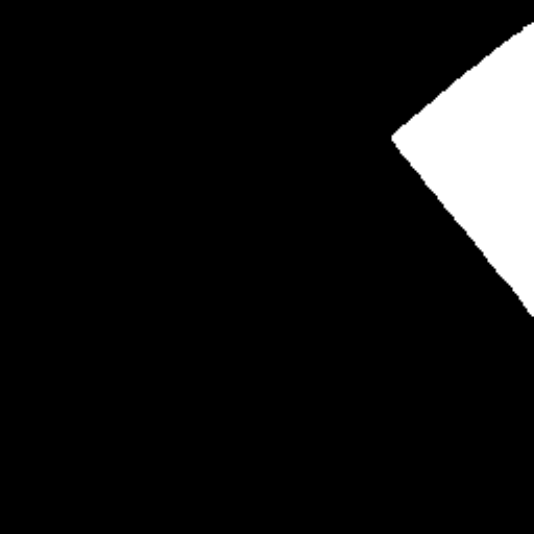} \end{minipage}                             \\
                                                        & \textbf{\textbf{River}}        
                                                                                        & \begin{minipage}{.1\textwidth} \includegraphics[width=0.75in]{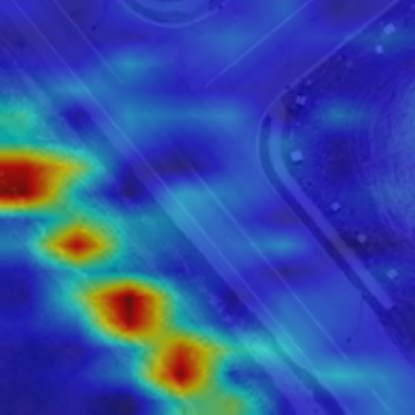} \end{minipage}                                    
                                                                                        & \begin{minipage}{.1\textwidth} \includegraphics[width=0.75in]{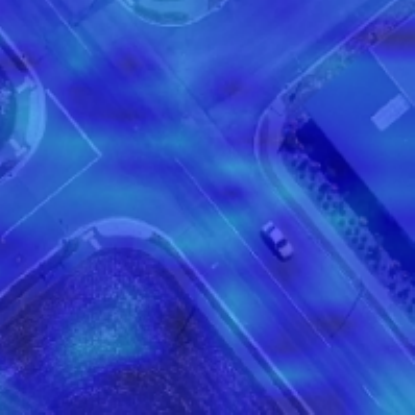} \end{minipage}                                    
                                                                                        & \begin{minipage}{.1\textwidth} \includegraphics[width=0.75in]{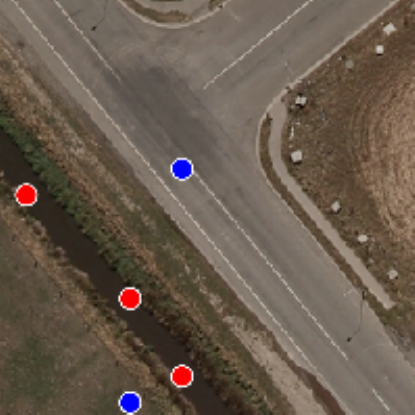} \end{minipage}                               
                                                                                        & \begin{minipage}{.1\textwidth} \includegraphics[width=0.75in]{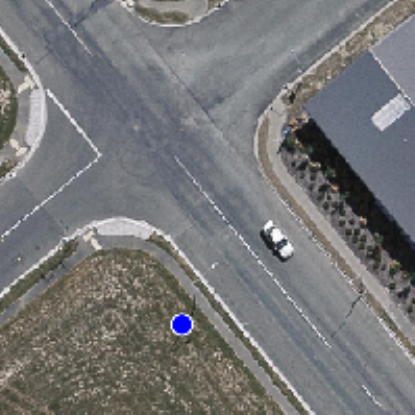} \end{minipage}                               
                                                                                        & \begin{minipage}{.1\textwidth} \includegraphics[width=0.75in]{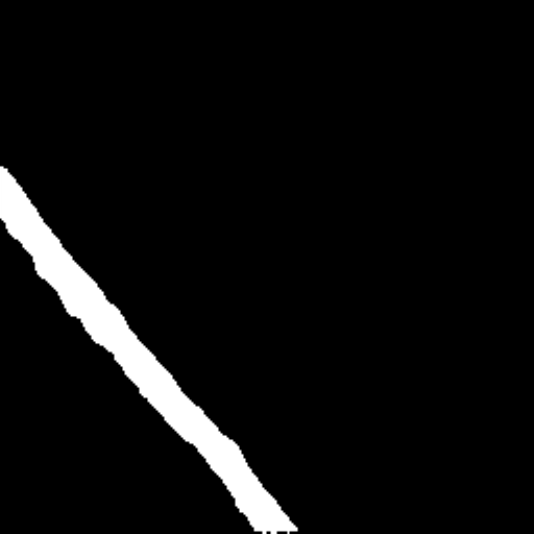} \end{minipage}                             \\
                                                        & \textbf{\textbf{Road}} 
                                                                                        & \begin{minipage}{.1\textwidth} \includegraphics[width=0.75in]{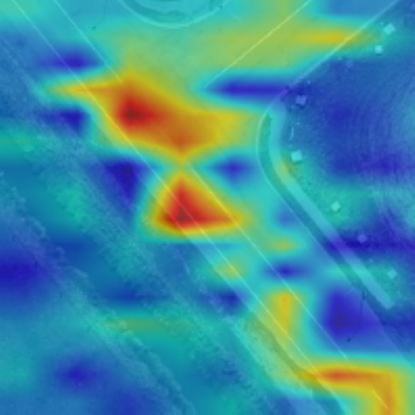} \end{minipage}                                    
                                                                                        & \begin{minipage}{.1\textwidth} \includegraphics[width=0.75in]{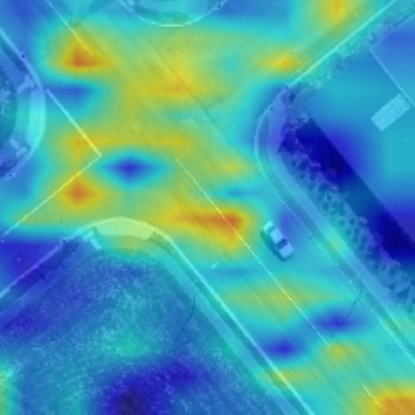} \end{minipage}                                    
                                                                                        & \begin{minipage}{.1\textwidth} \includegraphics[width=0.75in]{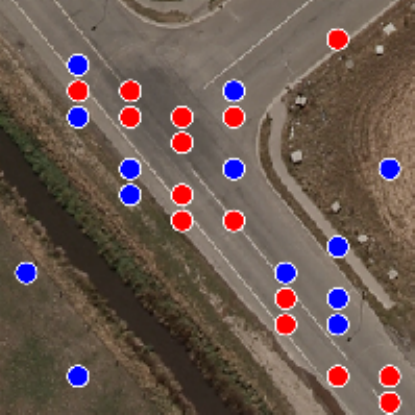} \end{minipage}                               
                                                                                        & \begin{minipage}{.1\textwidth} \includegraphics[width=0.75in]{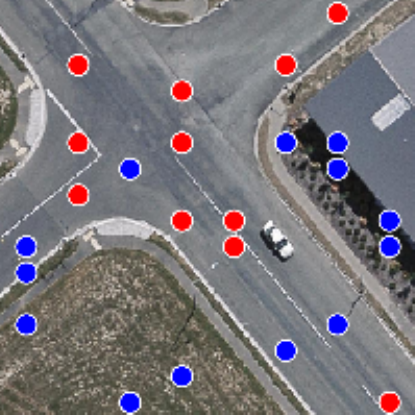} \end{minipage}                               
                                                                                        & \begin{minipage}{.1\textwidth} \includegraphics[width=0.75in]{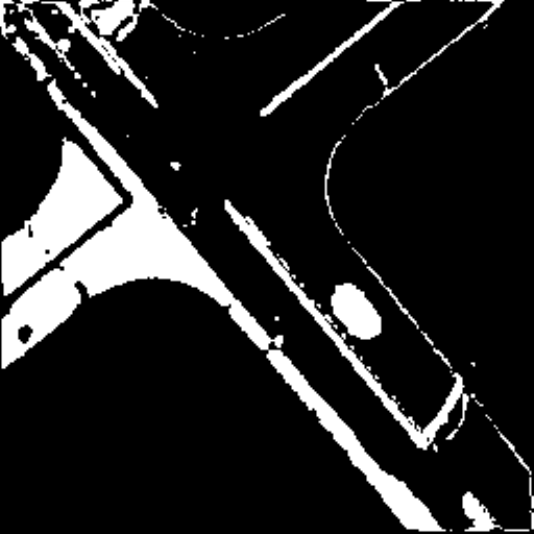} \end{minipage}                             
  \end{tblr}
  \end{table*}
As Table \ref{tab:table6} shows, for different CRoIs in the same pair of images, as long as the corresponding text description is given, 
the AUWCD can focus on the corresponding positions in the same pair of images, generate visual prompts, and finally obtain the corresponding CRoI CD results.
\subsection{Ablation Experiment}
This ablation experiment investigates the impact of VLMs in the semantics align module on the overall performance of the AUWCD. The study explored the following four aspects:

\paragraph{Effect of CRoI Text Prompts}
This experiment investigated the influence of different prompts on the accuracy of CRoI descriptions.
Since different prompts can lead to varying levels of precision in describing CRoIs, the objective is to identify prompts that provide more accurate descriptions of the target.

\paragraph{Effect of CRoUI Text Prompts}
In the Semantic Align Module, we can design a negative text prompt to suppress the model's recognition of CRoUI. 
The purpose of this experiment is to demonstrate that negative text prompts for CRoUIs can enhance CRoI detection quality by inhibiting CRoUI recognition.

\paragraph{Effect of Thresholds}
As thresholds directly affect the number and prompt point locations, determining the most suitable threshold values is essential.
The experiment aims to find optimal thresholds that maximize the CD process performance.

\paragraph{Impact of VLM Model Performance}
This analysis aims to demonstrate how VLM performance in the Semantic Align Module affects overall AUWCD performance.
The experimental details and results for each of the above aspects are presented below:
\subsubsection{Implementation Details}
We utilize the BCDD dataset as the data source for this experiment. 
Since the AUWCD does not require training and can directly perform CD tasks, we tested it on the entire BCDD dataset, which contains 7,434 pairs of images.
The implementation details for the three aforementioned experiments are as follows:

\paragraph{Effect of CRoI text prompts}
Considering that the BCDD dataset focuses on CD for buildings,
we selected seven text prompts to describe the CRoI as follows: 'buildings', 'houses', 'roofs', 'building roofs', 'house roofs', 'building house roofs', and 'iron building house roofs'.
Additionally, to demonstrate that only relevant text prompts are effective, we added other text descriptions unrelated to buildings such as "tree", "road", and "river".
To generate point prompts using CLIP surgery, we utilized the largest model size, CSVIT-L-14, with a threshold set to 0.80.
For the FSM module, we employed the SAM official demo, which also uses the largest model size, sam\_vit\_h.
Similarly, both CSVIT-L-14 and sam\_vit\_h also used pretrained model parameters.
We conducted a quantitative analysis using five evaluation metrics to assess the impact of different text prompts on the final CD results.
Furthermore, since the BCDD dataset contains annotations for buildings in both time frames, we demonstrated the influence of different text pages on building segmentation in dual temporal images.
This presentation adds further persuasiveness to the results.

\paragraph{Effect of CRoUI Text Prompts}
Since this is a building CD task, CRoI refers to buildings. Therefore, we set the CRoI text prompt to "iron building house roofs".
In building CD, common pseudo changes include shadows, roads and so on. 
Here, we selected common pseudo changes as the CRoUI textual prompts, which are "road", "shadow", and "tree".
Additionally, to explore the combined effect of multiple CRoUIs, we set "road, shadow, tree" as the text prompt for multiple CRoUI combinations. 
We compared these CRoUI text prompts above with an empty textual prompt(" ") to demonstrate the effectiveness of this method.
The VLM in the Semantic Align module is CSVIT-L-14, with a threshold set to 0.80. The FSM in the CRoI Segment Module is sam\_vit\_h. Both of them used pretrained model weights.

\paragraph{Effect of thresholds}
Based on the results obtained from the previous experiment, where we identified the optimal text prompt for building CD,
we explored the impact of varying threshold values on CD and CRoI segmentation.
We investigated threshold values in the range of 0.5 to 0.95 with a step size of 0.05 for their effect on performance.
Again, we used CLIP surgery with the model size CSVIT-L-14 and the FSM module with the sam\_vit\_h model while using the best-performing text prompt from the first experiment.
CSVIT-L-14 and sam\_vit\_h used pretrained model weights.

\paragraph{Impact of VLM performance}
This experiment explored the impact of VLM performance on the overall results.
Larger model sizes have been shown to improve performance
\cite{emergent_abilities_of_large_language_models, clip};
thus, we chose three different CLIP surgery model sizes to represent VLMs with varying capabilities.
The selected model sizes, from smallest to largest, were CSVIT-B-32, CSVIT-B-16, and CSVIT-L-14. 
All CLIP surgery models used pretrained model weights.
The text prompt remained consistent, using the best-performing prompt obtained from the previous experiment.
The threshold was set to the optimal value identified in the threshold experiment.
By conducting these three experiments, we obtained comprehensive insights into the effects of text prompts, thresholds, and VLM model sizes on the performance of the AUWCD using the BCDD dataset.
The quantitative analysis, combined with the building segmentation results, provided a more convincing evaluation of the proposed approach.

\subsubsection{Result}

\paragraph{Ablation experimental results of CRoI text prompts} 
The building CD indicators for the different CRoI text prompts in the BCDD dataset are shown in Table\ref{tab:table7}.\par
\begin{table}
  \caption{Building CD indicators for different CRoI text prompts in the BCDD dataset.\label{tab:table7}}
  \centering
  \begin{tabular}{cccccc}
    \begin{tblr}{
      cells = {c},
      hline{1-2,5,12} = {-}{},
    }
  \textbf{CRoI Text}                        & \textbf{PAcc}   & \textbf{F1}     & \textbf{Rec.} & \textbf{Pre.} & \textbf{OA}      \\
  
  \textbf{"tree"}                      & 42.58          & 11.79          & 44.90          & 6.79            & 70.91           \\
  \textbf{"road"}                      & 42.98          & 7.50          & 26.11          & 4.38            & 72.09           \\
  \textbf{"river"}                     & 42.96          & 10.76          & 45.36          & 6.10             & 67.39           \\
  
  \textbf{"buildings"}                 & 64.40          & 17.86          & 67.97          & 10.28             & 72.92           \\
  \textbf{"houses"}                    & 64.92          & 17.28          & 66.66          & 9.92              & 72.34           \\
  \textbf{"roofs"}                     & 70.17          & 18.66          & 70.91          & 10.75             & 73.22           \\
  \textbf{"building roofs"}            & 71.15          & 19.57          & 71.82          & 11.33             & 74.43           \\
  \textbf{"house roofs"}               & 71.60          & 19.13          & 71.84          & 11.03             & 73.69           \\
  \textbf{"building house roofs"}      & 72.07          & 19.69          & 72.18          & 11.40             & 74.49           \\
  \textbf{"iron building house roofs"} & \textbf{74.07} & \textbf{20.72} & \textbf{75.16} & \textbf{12.01}    & \textbf{75.08} 
  \end{tblr}
  \end{tabular}
  \end{table}
The building extraction indicators of the different CRoI text prompts in the BCDD dataset are shown in Table\ref{tab:table8}. \par

\begin{table}
  \caption{Building extraction indicators for different CRoI text prompts in the BCDD dataset.\label{tab:table8}}
  \centering
  \begin{tabular}{cccccc}
    \begin{tblr}{
      cells = {c},
      hline{1-2,5,12} = {-}{},
    }
  \textbf{CRoI Text}                        & \textbf{PAcc}   & \textbf{F1}     & \textbf{Rec.} & \textbf{Pre.} & \textbf{OA}      \\
  
  \textbf{"tree"}                           & 42.58           & 8.87            & 10.00         & 8.00             & 65.03           \\
  \textbf{"road"}                           & 42.98           & 7.32            & 8.21          & 6.60             & 64.50           \\
  \textbf{"river"}                          & 42.96           & 11.94           & 14.18         & 10.32            & 64.30           \\
  \textbf{"buildings"}                      & 64.40           & 36.17          & 44.86          & 30.30             & 72.96           \\
  \textbf{"houses"}                         & 64.92           & 34.44          & 41.63          & 29.36             & 72.93           \\
  \textbf{"roofs"}                          & 70.17           & 38.68          & 49.77          & 31.63             & 73.06           \\
  \textbf{"building roofs"}                 & 71.15           & 39.95          & 50.82          & 32.91             & 73.91           \\
  \textbf{"house roofs"}                    & 71.60           & 39.24          & 50.18          & 32.22             & 73.47           \\
  \textbf{"building house roofs"}           & 72.07           & 40.01          & 50.00          & 33.35             & 74.40           \\
  \textbf{"iron building house roofs"}      & \textbf{74.07}  & \textbf{42.81} & \textbf{54.50} & \textbf{35.24}    & \textbf{75.13} 
  \end{tblr}
  \end{tabular}
  \end{table}
According to the experimental results in the table above, the text prompt was "iron building house roofs".
Regardless of the building CD task or the building extraction task, the five evaluation indicators were better than other text prompts.
It can be concluded that "iron building house roofs" provide a better description of the building used in CLIP surgery; therefore, these roofs were used as a text prompt for later ablation experiments. \par

In addition, the experimental results obtained by different text prompts have considerable differences in terms of the indicators.
In the building CD and extraction tasks, using text descriptions unrelated to buildings leads to a significant decline in the model's detection performance,
which further demonstrates the effectiveness of our Semantic Align Module. 
Additionally, among the various textual descriptions for buildings, the maximum differences in F1 scores can reach 3.44 and 8.37 percentage points, respectively.
In fact, there are so many text prompts that the prompt "iron building house roofs" may not be the most effective text prompt.
Considerable research has shown that the zero-shot ability of large models largely depends on the quality of the prompts. This article provides prompt words according to the categories corresponding to the CRoI.\par

Therefore, providing higher-quality and more stable CRoI semantic prompts for VLMs can also be an important direction for optimizing the AUWCD in the future. \par

\paragraph{Ablation experimental results of CRoUI text prompts} 

The building CD indicators of the different CRoUI text prompts in the BCDD dataset are shown in Table\ref{tab:table9}. 
The Experimental results show that using common pseudo-change text prompts in building CD tasks as CRoUI leads to improvements across four metrics compared to using empty text prompt(" "). 
Therefore, this method indeed enhances the overall detection performance of the model by suppressing the recognition of CRoUI. 
Additionally, combining multiple CRoUI text prompts demonstrated better performance than using a single CRoUI.
This indicates that this method can better optimize the model's detection performance by combining multiple CRoUIs.
\par

\begin{table}
  \caption{
    Building CD indicators of different CRoUI text prompts in the BCDD dataset.\label{tab:table9}
    }
  \centering
  \begin{tabular}{ccccc}
    \begin{tblr}{
      cells = {c},
      hline{1-3,6-7} = {-}{},
    }
  \textbf{CRoUI Text}           & \textbf{F1}                                                                               & \textbf{Rec.}                                                                             & \textbf{Pre.}                                                                           & \textbf{OA}      \\
  \textbf{" "}                   &  20.72                                                                                    & 75.16                                                                                       & 12.01                                                                                       & 75.08           \\
  \textbf{"road"}               & \begin{tabular}[c]{@{}c@{}}23.00\\(\textcolor{red}{+2.28})\end{tabular}                   & \begin{tabular}[c]{@{}c@{}}80.32\\(\textcolor{red}{+5.16})\end{tabular}                     & \begin{tabular}[c]{@{}c@{}}13.40\\(\textcolor{red}{+1.39})\end{tabular}                   & \begin{tabular}[c]{@{}c@{}}76.66\\(\textcolor{red}{+1.58})\end{tabular}           \\
  \textbf{"shadow"}             & \begin{tabular}[c]{@{}c@{}}21.37\\(\textcolor{red}{+0.65})\end{tabular}                   & \begin{tabular}[c]{@{}c@{}}76.76\\(\textcolor{red}{+1.60})\end{tabular}                     & \begin{tabular}[c]{@{}c@{}}12.41\\(\textcolor{red}{+0.40})\end{tabular}                   & \begin{tabular}[c]{@{}c@{}}75.53\\(\textcolor{red}{+0.45})\end{tabular}           \\
  \textbf{"tree"}               & \begin{tabular}[c]{@{}c@{}}23.35\\(\textcolor{red}{+2.63})\end{tabular}                   & \begin{tabular}[c]{@{}c@{}}\textbf{80.80}\\(\textbf{\textcolor{red}{+5.64}})\end{tabular} & \begin{tabular}[c]{@{}c@{}}13.65\\(\textcolor{red}{+1.64})\end{tabular}                   & \begin{tabular}[c]{@{}c@{}}77.03\\(\textcolor{red}{+1.95})\end{tabular}           \\
  \textbf{"road, shadow, tree"} & \begin{tabular}[c]{@{}c@{}}\textbf{23.38}\\(\textbf{\textcolor{red}{+2.66}})\end{tabular} & \begin{tabular}[c]{@{}c@{}}79.57\\(\textcolor{red}{+4.41})\end{tabular}                     & \begin{tabular}[c]{@{}c@{}}\textbf{13.71}\\(\textbf{\textcolor{red}{+1.70}})\end{tabular} & \begin{tabular}[c]{@{}c@{}}\textbf{77.41}\\(\textbf{\textcolor{red}{+2.33}})\end{tabular} 
  \end{tblr}
  \end{tabular}
  \end{table}

\paragraph{Ablation experimental results of thresholds} 
From the text prompt ablation experiment in the previous section, we obtained the prompt "iron building house roofs", which better describes the building.
In the threshold ablation experiment in this section, we use this as a fixed prompt word.\par
\begin{figure}[htbp]
  \centering
  \includegraphics[width=3.6in]{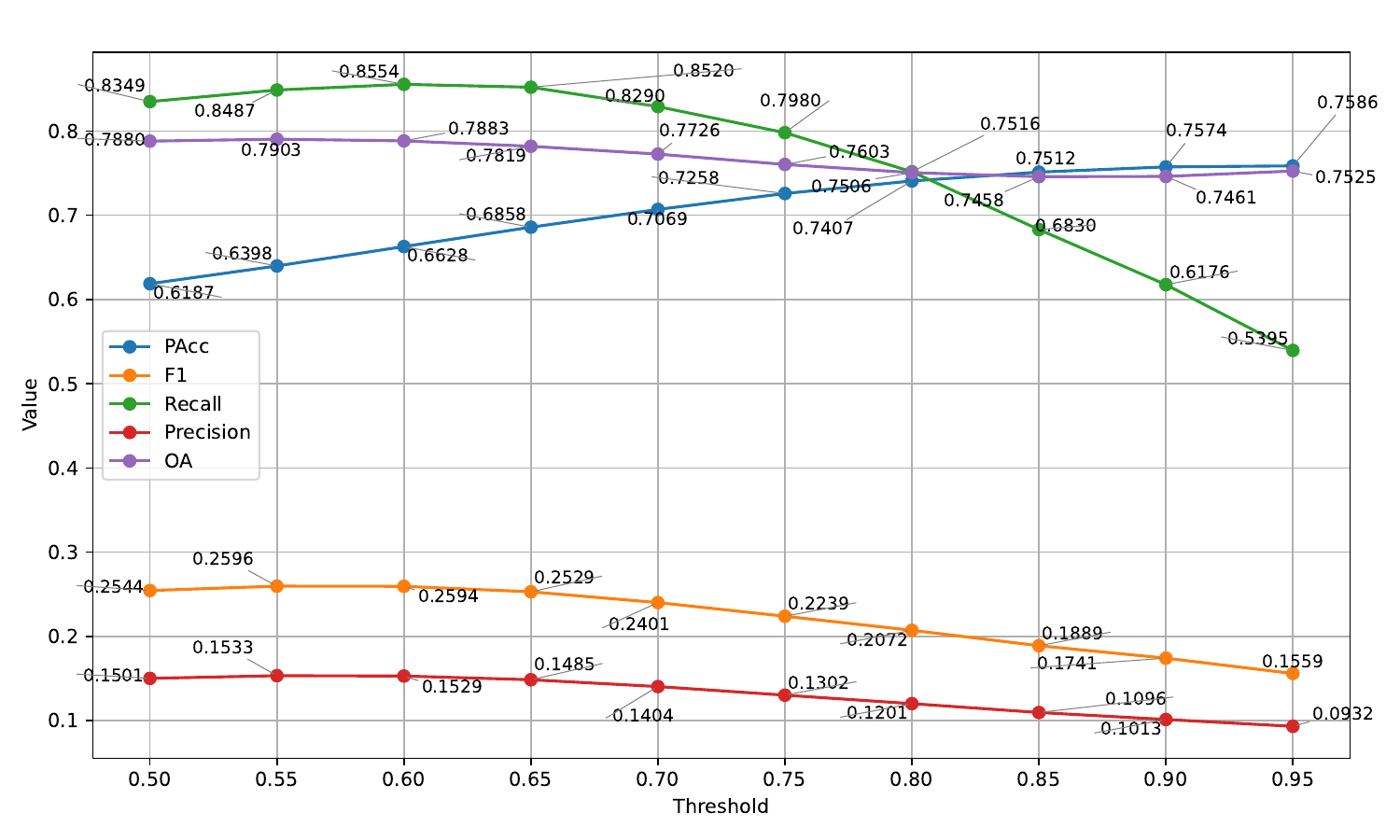}
  \caption{Different thresholds for building CD indicators in the BCDD dataset.}
  \label{fig9}
  \end{figure}

The evaluation results of different thresholds for building CD in the BCDD dataset are shown in Figure \ref{fig9}.\par

\begin{figure}[htbp]
    \centering
    \includegraphics[width=3.6in]{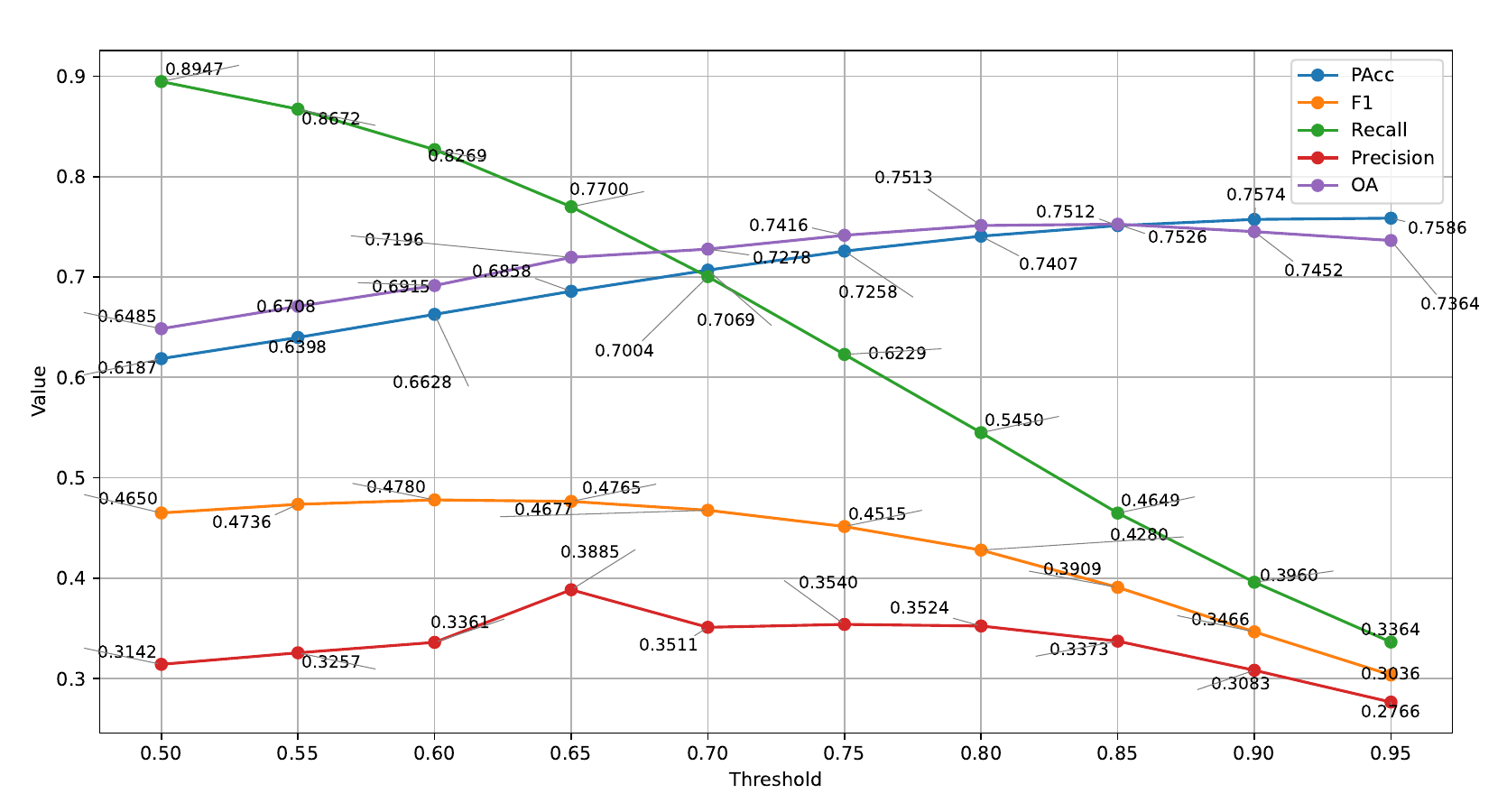}
    \caption{Different thresholds for building extraction indicators in the BCDD dataset.}
    \label{fig10}
    \end{figure}

The evaluation results of different thresholds for building extraction in the BCDD dataset are shown in Figure \ref{fig10}.\par

With respect to the F1 score, the most important indicator on the CD task, when the threshold is 0.55,
the F1 score on the building CD task of the BCDD dataset reaches the highest level of 0.2596, and the F1 score on the building extraction task reaches 0.4736, which is almost the highest level. 
Therefore, when the threshold is 0.55, the visual prompt points generated by CLIP surgery are better than those generated by the other methods; therefore, 0.55 was used as the threshold for subsequent ablation experiments.\par

The experimental results in the above figure show that as the threshold changes, various indicators also exhibit some regular changes.
For example, as the threshold gradually increases, the recall shows an obvious negative correlation.
This is because, under a high similarity selection threshold, ignoring areas in the image that are less similar to CRoI text descriptions is easy, resulting in missed CRoI detections.
The PAcc shows an obvious positive correlation with the threshold.
The number of visual prompt points under high similarity conditions can be accurately increased, indicating that the model is more focused on real CRoIs.
Moreover, the change patterns of the F1 score and precision index show a parabolic pattern that first increases and then decreases.
However, there is great uncertainty in finding the parabolic vertex threshold, which requires more thorough research and more appropriate methods. \par

In addition, the CD error is significantly larger than the extraction error for the building CD and building extraction evaluation indicators.
We performed the following analysis on this phenomenon. 
For the building CD and building extraction indicators under different thresholds, we used $\mathit{F1_{CD}}$ and $\mathit{F1_{Seg}}$ to represent the F1 scores of building CD and extraction, respectively, and
$\mathit{1-F1_{CD}}$ and  $\mathit{1-F1_{Seg}}$ to represent their errors, respectively. The error ratio can be calculated using the following formula (\ref{eq18}):\par

\begin{equation}
  \label{eq18}
  \mathit{Error Ratio}=\mathit{\frac{1-F1_{CD}}{1-F1_{Seg}} }  
\end{equation}

\begin{figure}[htbp]
  \centering
  \includegraphics[width=3.4in]{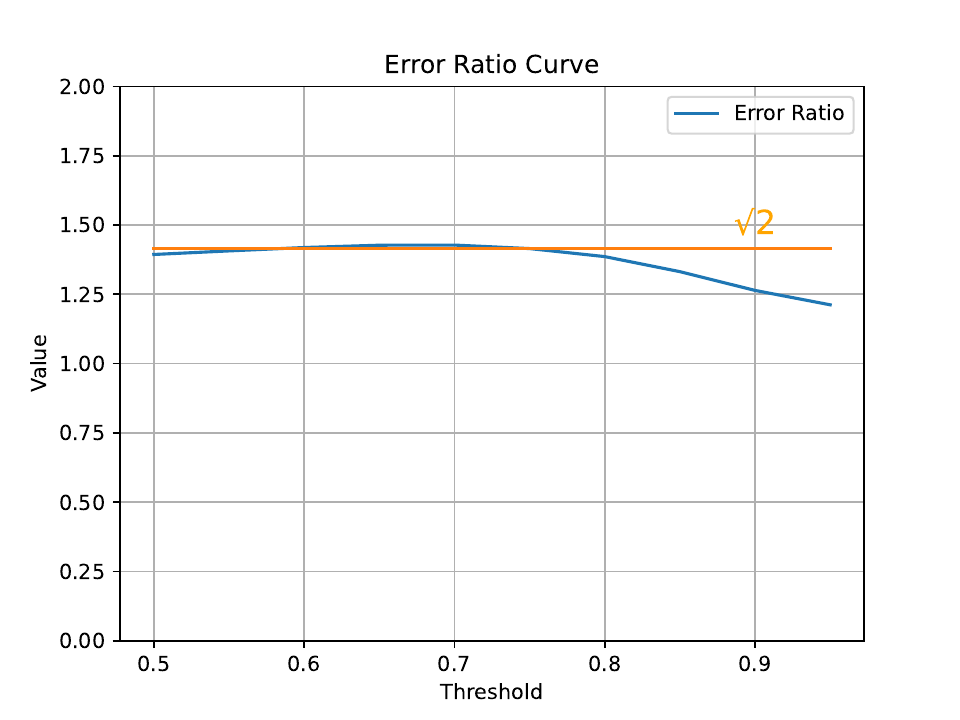}
  \caption{Error Ratio Curve.}
  \label{fig11}
\end{figure}

This error ratio curves under different thresholds are shown in Fig.\ref{fig11}: \par

Fig.\ref{fig11} shows that the error ratio of the CD score to the extracted F1 score of the building remains approximately $\sqrt{2}$.
This error accumulation occurs because the method for obtaining the change map in the experiment (i.e., the change process method in the CD module) involves comparing two masks (operations such as union, intersection, and complement).
Therefore, the error of any mask may have an impact on the final result. In statistics and error analysis, when independent error sources are combined, the square of the total error is the sum of the squares of the independent errors, that is:\par

\begin{equation}
  \label{eq19}
  \mathit{M^2}=\mathit{m_{1}^{2}+m_{2}^{2}+\cdots +m_{n}^{2}}
\end{equation}

where $\mathit{M}$ represents the independent errors that constitute the total error.
The CRoI extraction mask errors of the two phases are assumed to be independently and identically distributed, and the size of each mask error is assumed to be m.
Therefore, when the change map is obtained by merging the CRoI extraction masks of the two phases, according to formula (\ref{eq19}), the square of the change map error
$\mathit{m_{c}^{2}=m^2+m^2}=2m^2$; therefore, $\mathit{m_{c}}=\sqrt{2}m$.
In complex systems, error propagation may nonlinearly affect the final output, which is why the actual error ratio curve does not coincide with $\sqrt{2}$.
Therefore, one of the views of this article on the ratio of the error ratio close to $\sqrt{2}$ is that for obtaining the change map using only the CRoI extraction masks in the two phases,
the CRoI extraction errors of the two phases are independent are independent to some extent.
The total error contributes similarly.
Therefore, changing the change process method in the CD module in the AUWCD is proposed to reduce the error accumulation caused by error propagation as a future research direction.\par

\paragraph{Ablation experimental results of VLM performance} 
The test prompt and threshold selected in this section of the experiment are the optimal results obtained in the above ablation experiment, which are "iron building house roofs" and 0.55, respectively.
CLIP surgery of different scales are selected to represent VLMs with different performances.\par
The experimental results of building extraction on the BCDD dataset are shown in Table \ref{tab:table10}.\par
\begin{table*}
  \caption{Building CD indicators of CLIP surgery models of different scales on the BCDD dataset\label{tab:table10}}
  \centering
  \begin{tabular}{ccccccc}
  \begin{tblr}{
    cells = {c},
    hline{1-2,5} = {-}{},
  }
  \textbf{Model}      & \textbf{Size}    & \textbf{PAcc}   & \textbf{F1}     & \textbf{Recall} & \textbf{Precision} & \textbf{OA}      \\
  \textbf{CSVIT-B-16} & 350.8MB          & 0.5464          & 0.1542          & 0.5975          & 0.0884             & 0.7159           \\
  \textbf{CSVIT-B-32} & 354MB            & 0.5914          & 0.1863          & 0.7665          & 0.1060             & 0.7099           \\
  \textbf{CSVIT-L-14} & \textbf{932.8MB} & \textbf{0.6398} & \textbf{0.2596} & \textbf{0.8487} & \textbf{0.1533}    & \textbf{0.7903} 
  \end{tblr}
  \end{tabular}
  \end{table*}

The experimental results of building extraction on the BCDD dataset are shown in Table \ref{tab:table11}.\par
\begin{table*}
  \caption{Building extraction indicators of CLIP surgery models of different scales on the BCDD dataset\label{tab:table11}}
  \centering
  \begin{tabular}{ccccccc}
  \begin{tblr}{
      cells = {c},
      hline{1-2,5} = {-}{},
    }
  \textbf{Model}      & \textbf{Size}    & \textbf{PAcc}   & \textbf{F1}     & \textbf{Recall} & \textbf{Precision} & \textbf{OA}      \\
  \textbf{CSVIT-B-16} & 350.8MB          & 0.5464          & 0.3735          & 0.8156          & 0.2422             & 0.5329           \\
  \textbf{CSVIT-B-32} & 354MB            & 0.5914          & 0.4178          & 0.6516          & 0.3075             & \textbf{0.6899}  \\
  \textbf{CSVIT-L-14} & \textbf{932.8MB} & \textbf{0.6398} & \textbf{0.4736} & \textbf{0.8672} & \textbf{0.3257}    & 0.6708          
  \end{tblr}
  \end{tabular}
  \end{table*}

The experimental results reveal that CSVIT-L-14, the largest model, outperforms the other smaller models in both building CD and extraction tasks.
As the model size continues to increase, the performance on detection and extraction tasks also improves.
As described in the previous implementation section, an increase in the model size improves the performance.
Therefore, this experiment proves that the CD performance of the AUWCD is positively correlated with the VLM performance.\par

On the basis of this conclusion, we propose that an increase in model size leads to stronger language and image representation and understanding capabilities,
thereby enabling a better understanding of CRoI text descriptions and their corresponding image representations.
However, research is needed to explore this hypothesis. In addition, only CLIP surgery was used for ablation experiments in this article.
In fact, the performance of CLIP surgery is not ideal in some scenarios, especially in remote sensing
\cite{clip}.
Therefore, exploring VLMs with better remote sensing performance is also an important direction for optimizing the CD effect of AUWCDs in the future. \par

\section{Conclusion and Future Work}
In this article, we first discussed the nature of the CD task and proposed the view that "semantics is the first imaging factor of the CD task".
Moreover, the current CD method relies on the vision-first ViFi-CD paradigm; however, a serious problem with this paradigm is that it inevitably ignores some potential CRoIs and has difficulty adapting to the CD tasks of different CRoIs.
To solve this problem, a "semantic-first" CD paradigm called SeFi-CD was proposed based on the above. To verify the effectiveness of this paradigm, this paper proposes a new CD framework AUWCD under this paradigm.
Compared with state-of-the-art methods in the current ViFi-CD paradigm, AUWCD performs well on both the SECOND and BCDD datasets, outperforming the state-of-the-art current CD methods when dynamically switching CRoIs without retraining.
These findings verify the effectiveness and necessity of the paradigm shift proposed in this article.\par

In the future, we will conduct further research on more complex CD tasks and simultaneously build a "semantic-first" paradigm CD dataset for optimizing and evaluating subsequent methods to promote further development of this paradigm.\par

\bibliography{IEEEabrv,tgrs}
\begin{IEEEbiography}
[{\includegraphics[width=1in,height=1.25in,clip,keepaspectratio]{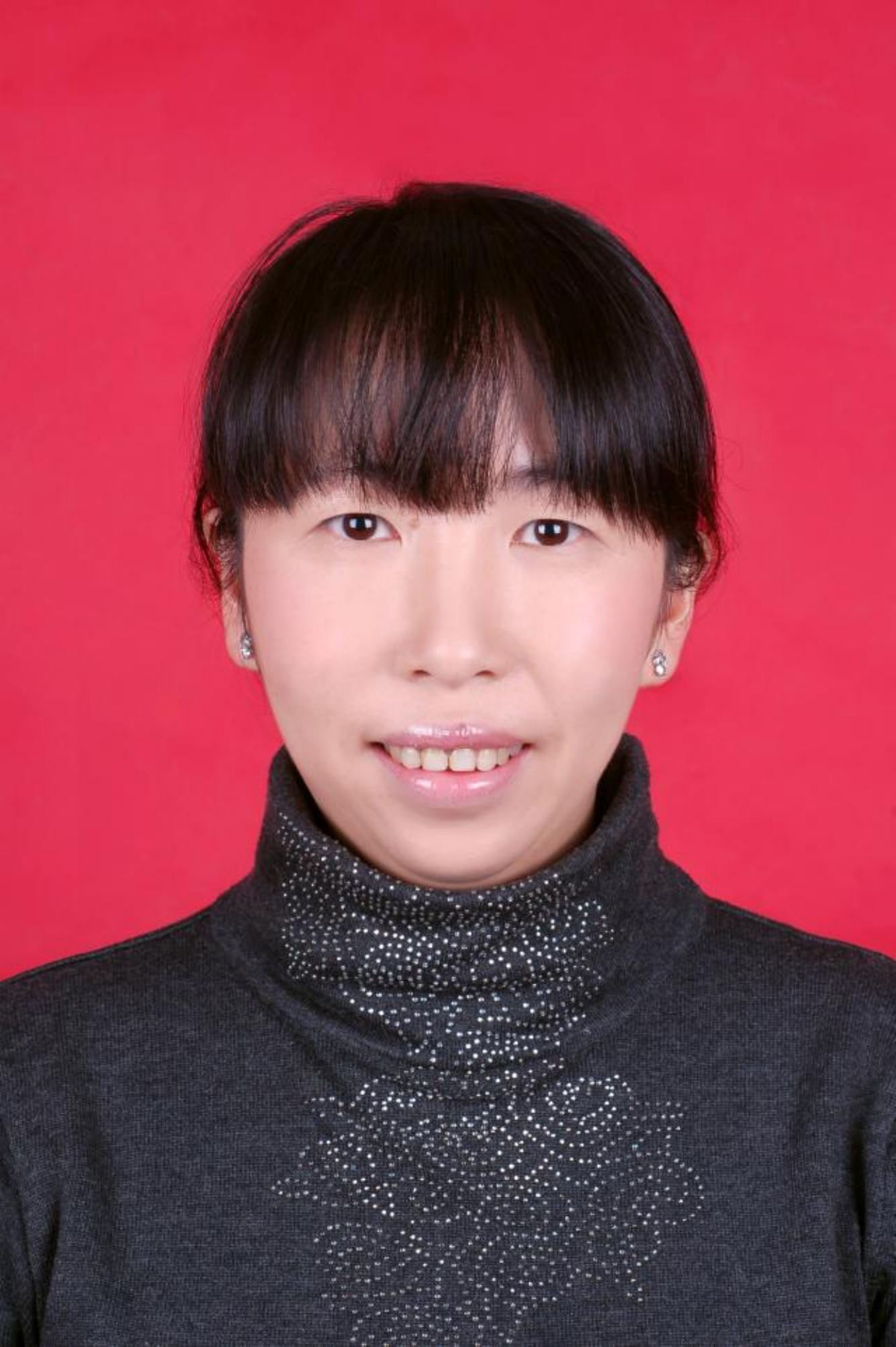}}]
{Lin Zhao} obtained her Ph.D. degree in School of Geosciences and Info-Physics, Central South University.
She is currently an associate professor with the School of Geosciences and Info-Physics, Central South University, Changsha, China.
Her research interests mainly concentrate in humanities and social science based on big geodata and artificial intelligence.
She has published more than 10 peer-reviewed journal articles.
\end{IEEEbiography}

\begin{IEEEbiography}
[{\includegraphics[width=1in,height=1.25in,clip,keepaspectratio]{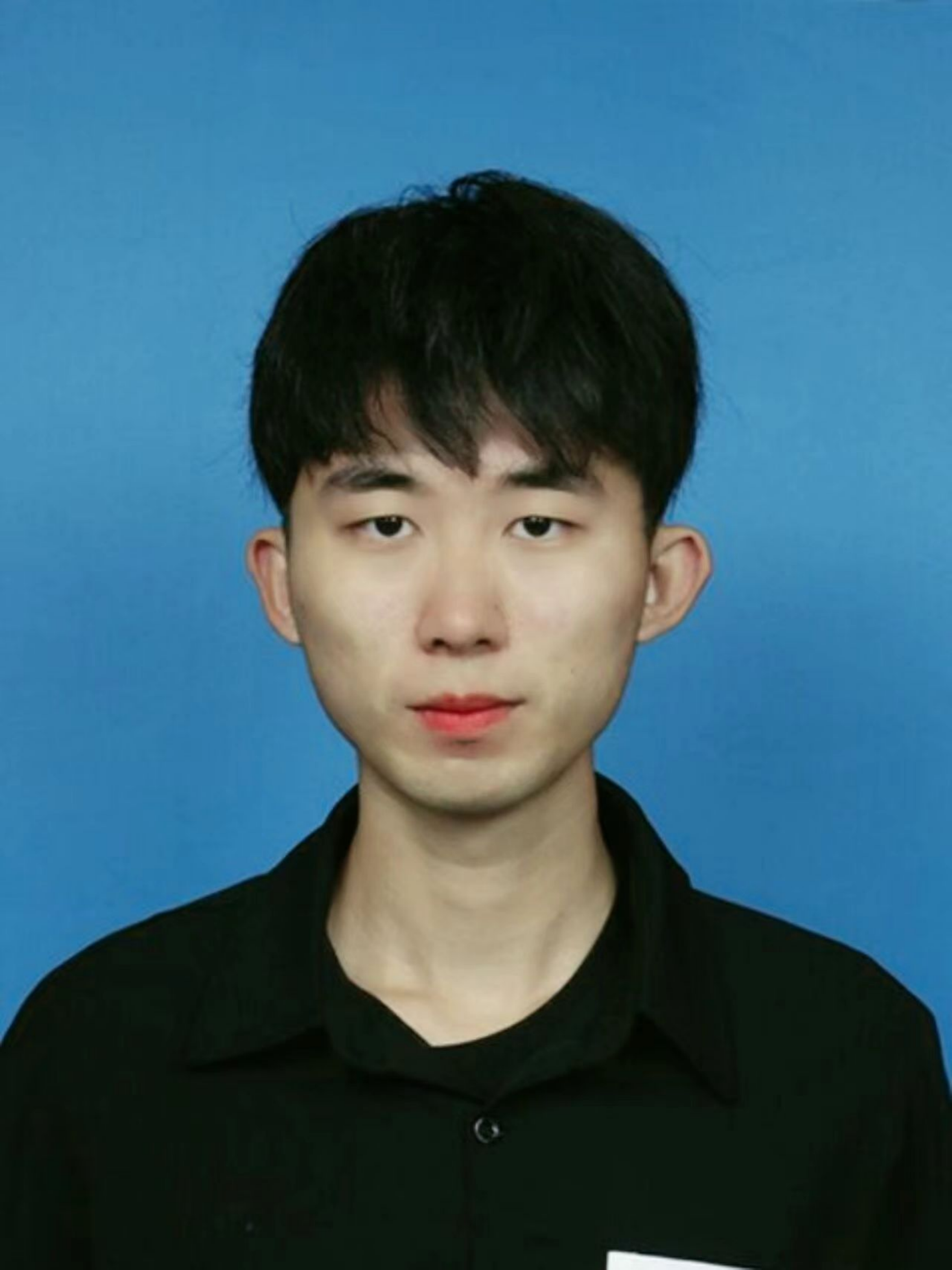}}]
{Zhenyang Huang} received the B.S. 
degree from the Shandong University of Science and Technology, Qing dao, China, in 2022.
He is pursuing the master's degree with the School of Geosciences and Info-Physics, Central South University, Changsha, China.
His research interests include computer vision, deep learning, multimodal large model, and remote sensing image change detection.
\end{IEEEbiography}

\begin{IEEEbiography}
[{\includegraphics[width=1in,height=1.25in,clip,keepaspectratio]{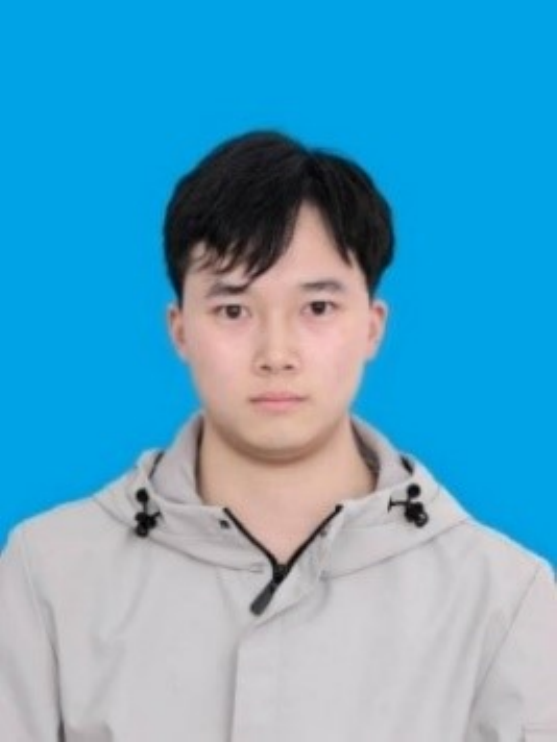}}]
{Dongsheng Kuang} received the B.S. degree from the Northwest University, Xi'an, China, in 2021.
He is pursuing the master's degree with the School of Geosciences and Info-Physics, Central South University, Changsha, China.
His research interests include computer vision, machine learning, deep learning, and remote sensing image change detection.
\end{IEEEbiography}

\begin{IEEEbiography}
[{\includegraphics[width=1in,height=1.25in,clip,keepaspectratio]{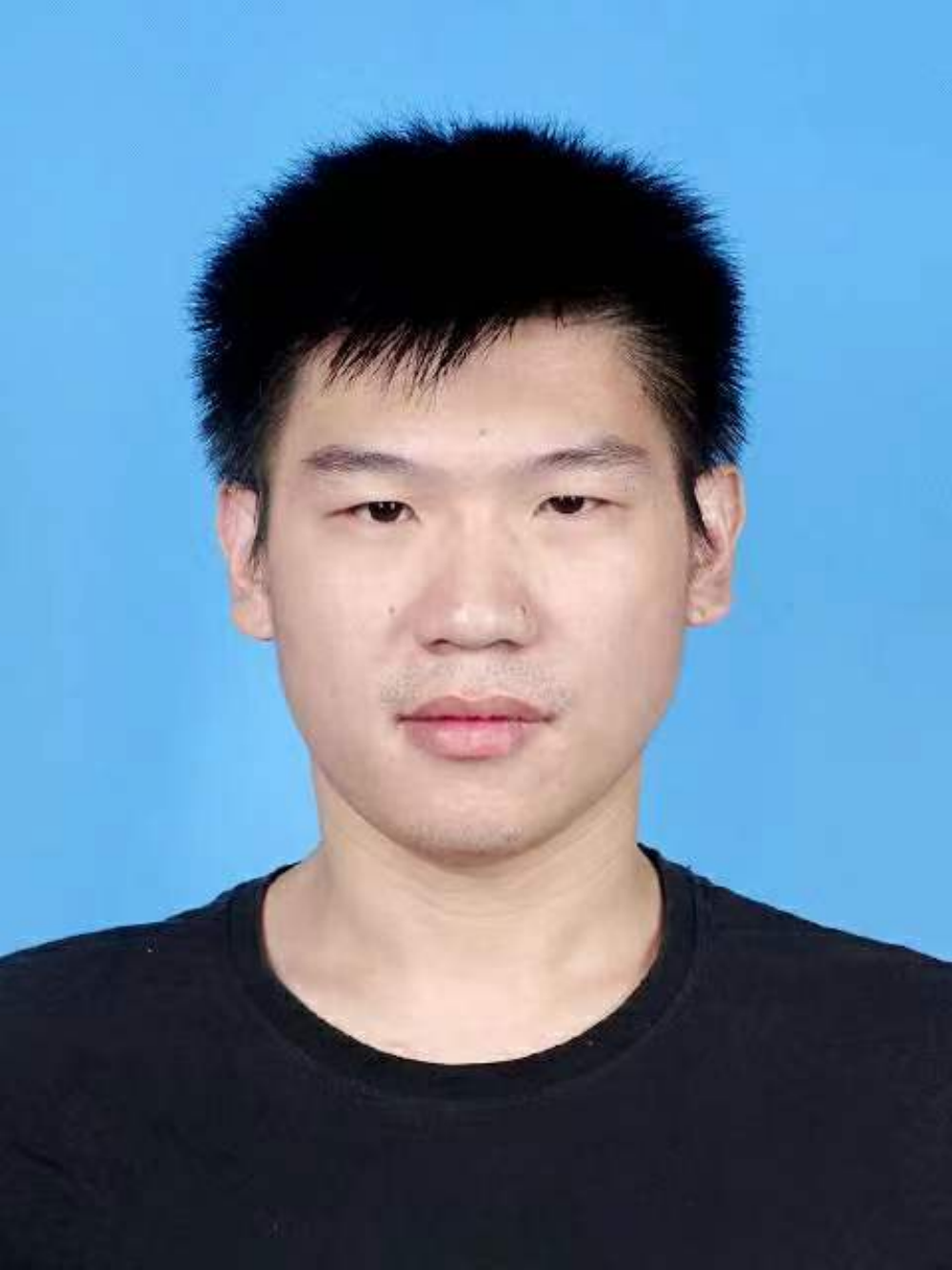}}]
{Chengli Peng} received the B.E. degree from the School of Electrical Engineering, Xinjiang University, Urumchi, China, in 2016, the M.S.
degree from the School of Engineering, Huazhong Agricultural University, Wuhan, China, in 2018, and the Ph.D. degree from the Electronic Information School, Wuhan University, Wuhan, China, in 2021.
He is currently a Research Associate with the School of Geosciences and Info-Physics, Central South University, Changsha, China.
His research interests include computer vision and deep learning.  
\end{IEEEbiography}

\begin{IEEEbiography}
[{\includegraphics[width=1in,height=1.25in,clip,keepaspectratio]{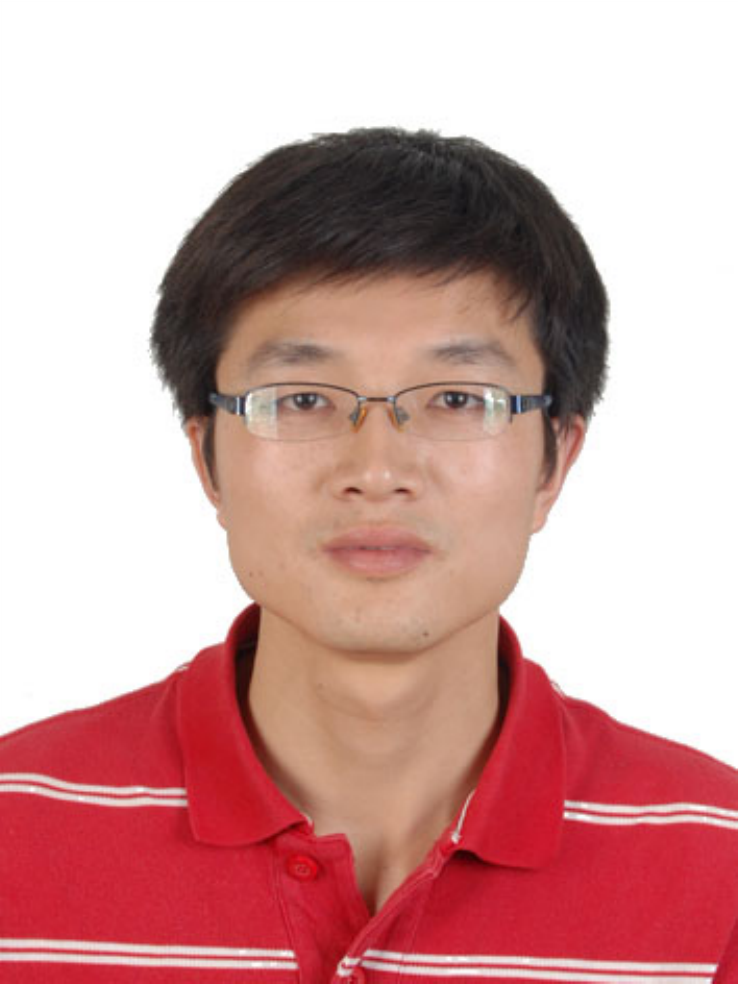}}]
{Jun Gan} received his master's degree in Surveying and Mapping Engineering from Wuhan University in 2009.
He is currently a professor of engineering at the Surveying, Mapping and Geoinformation Research Institute of China Railway Design Corporation.
His research interest includes aerial photogrammetry, satellite remote sensing technology, InSAR, and their application in railway engineering.
\end{IEEEbiography}

\begin{IEEEbiography}
[{\includegraphics[width=1in,height=1.25in,clip,keepaspectratio]{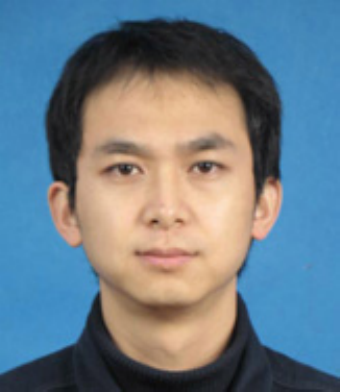}}]
{Haifeng Li} (M'15) received a master's degree in transportation engineering from the South China University of Technology, Guangzhou, China, in 2005, and a Ph.D. in photogrammetry and remote sensing from Wuhan University, Wuhan, China, in 2009.
He is currently a Professor with the School of Geosciences and Info-Physics, Central South University, Changsha, China.
He was a Research Associate with the Department of Land Surveying and Geo-Informatics, The Hong Kong Polytechnic University, Hong Kong, in 2011, and a Visiting Scholar with the University of Illinois at Urbana-Champaign, Urbana, IL, USA, from 2013 to 2014.
He has authored over 130 journal papers. 
His current research interests include geo/remote sensing big data, multi-modal large General Intelligence Model, Machine Learning/Deep Learning, and artificial/brain-inspired intelligence.
He is a reviewer for IEEE TNNLS, IEEE TIT, IEEE TGRS, IEEE TPAMI.
\end{IEEEbiography}

\end{document}